\documentclass[manuscript,acmsmall]{acmart}

\AtBeginDocument{%
  \providecommand\BibTeX{{%
    \normalfont B\kern-0.5em{\scshape i\kern-0.25em b}\kern-0.8em\TeX}}}

\setcopyright{acmlicensed}
\copyrightyear{2024}
\acmYear{2024}
\acmDOI{XXXXXXX.XXXXXXX}

\usepackage{enumitem}
\usepackage{booktabs}
\usepackage{siunitx}
\usepackage{tabularx}
\usepackage{booktabs} 
\usepackage{amsmath} 
\newcounter{rownum}
\usepackage{xcolor}

\newcommand{\rownumber}{\stepcounter{rownum}\textcolor{gray}{\arabic{rownum}}}

\begin{document}

\title{Dynamic Fairness Perceptions in Human-Robot Interaction}

\author{Houston Claure}
\affiliation{%
  \institution{Yale University}
  \country{USA}
}
\author{Kate Candon}
\affiliation{%
  \institution{Yale University}
  \country{USA}
}
\author{Inyoung Shin}
\affiliation{%
  \institution{Yale University}
  \country{USA}
}
\author{Marynel Vázquez}
\affiliation{%
  \institution{Yale University}
  \country{USA}
}

\renewcommand{\shortauthors}{Claure, et al.}

\begin{abstract}
People deeply care about how fairly they are treated by robots. The established paradigm for probing fairness in Human-Robot Interaction (HRI) involves measuring the perception of the fairness of a robot at the conclusion of an interaction. However, such an approach is limited as interactions vary over time, potentially causing changes in fairness perceptions as well. To validate this idea, we conducted a 2$\times$2 user study with a mixed design (N=40) where we investigated two factors: the timing of unfair robot actions (early or late in an interaction) and the beneficiary of those actions (either another robot or the participant). Our results show that fairness judgments are not static. They can shift based on the timing of unfair robot actions. Further, we explored using perceptions of three key factors (reduced welfare, conduct, and moral transgression) proposed by a Fairness Theory from Organizational Justice to predict momentary perceptions of fairness in our study. Interestingly, we found that the reduced welfare and moral transgression factors were better predictors than all factors together. Our findings reinforce the idea that unfair robot behavior can shape perceptions of group dynamics and trust towards a robot and pave the path to future research directions on moment-to-moment fairness perceptions.
\end{abstract}



\begin{teaserfigure}
\centering
\includegraphics[width=.49\linewidth]{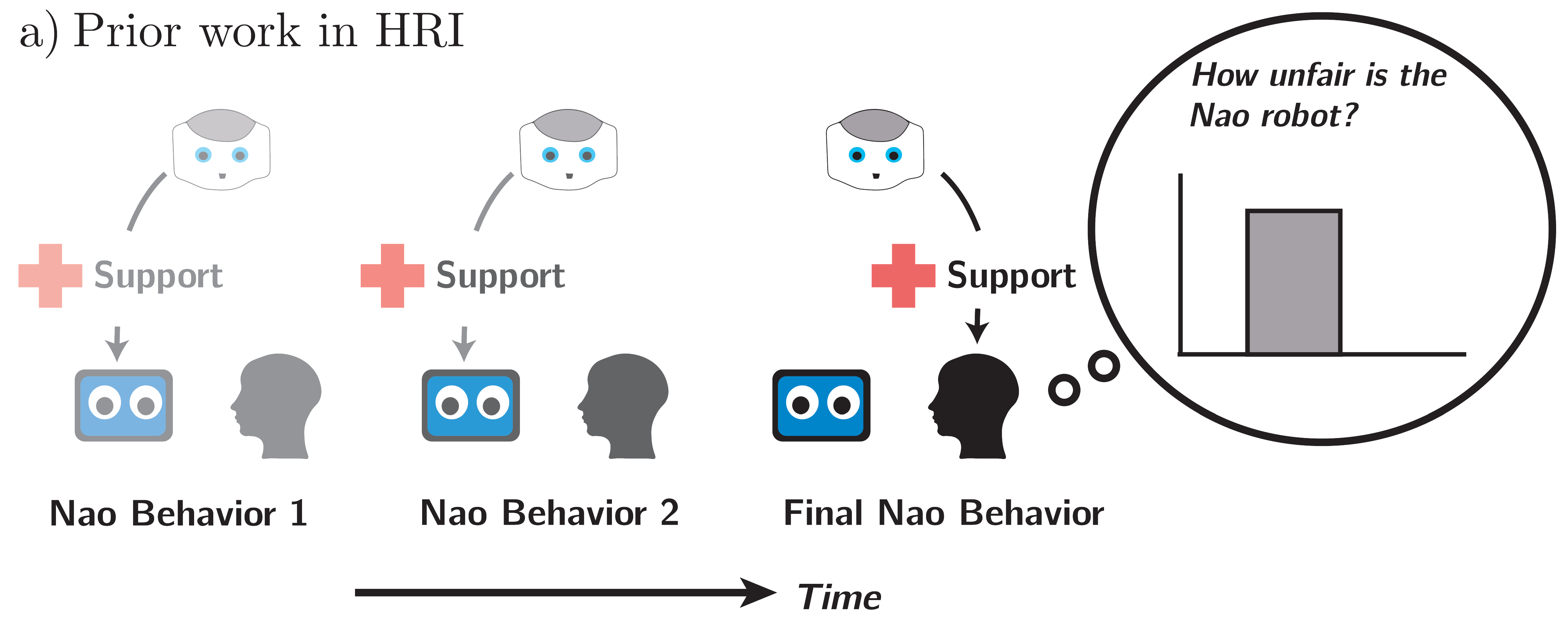}\hfill
\includegraphics[width=.49\linewidth]{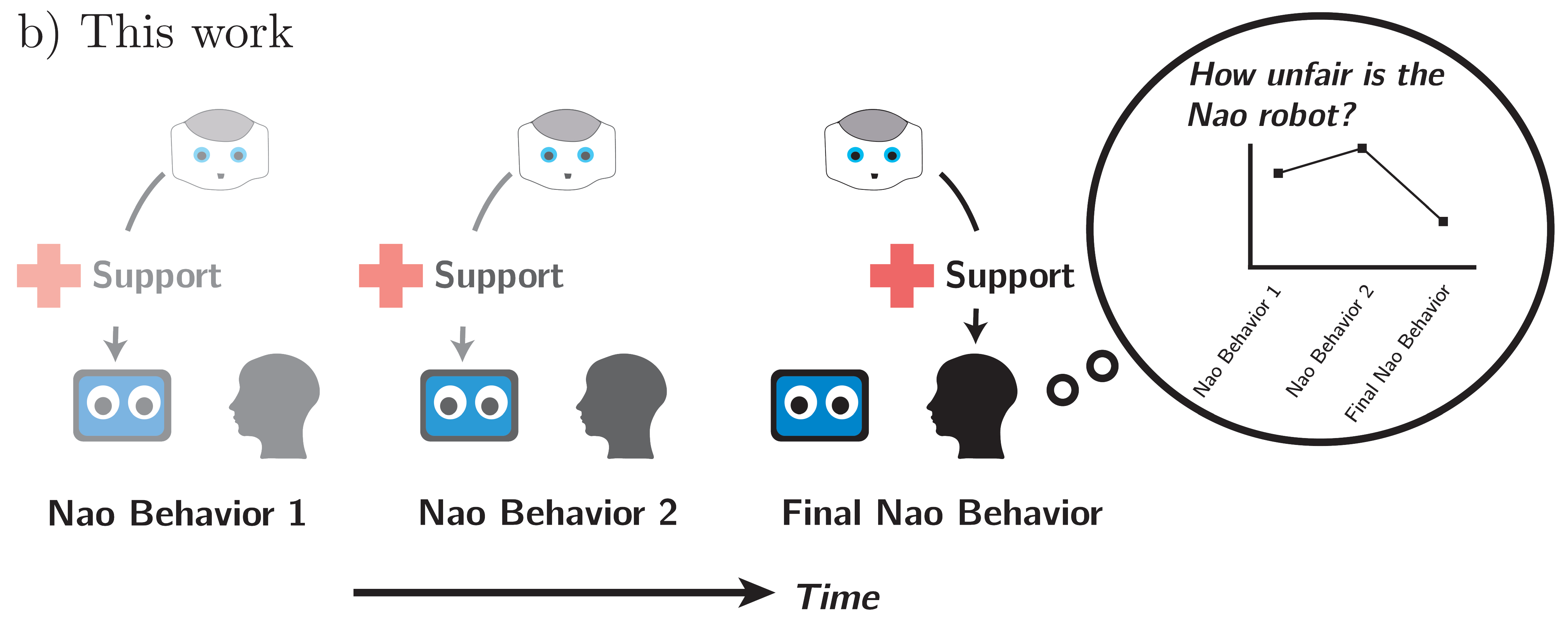}
\caption{How would humans perceive the unfair behavior of a Nao robot (white) that disproportionately supports another robot (blue) throughout an interaction? Prior work in Human-Robot Interaction has focused on measuring perceptions of fairness from humans once at the conclusion of social encounters with robots (a). In order to gain a deeper understanding of fairness perceptions, our work explores measuring these perceptions multiple times during human-robot interactions (b).}
\Description{Prior work in HRI has focused on measuring perceptions of fairness from humans at the conclusion of human-robot interactions (a). In order to gain a deeper understanding of fairness perceptions, our work explores measuring these perceptions multiple times during human-robot interactions (b).}
\label{fig:overview}
\end{teaserfigure}




\maketitle

\section{Introduction}
Fairness is a value deeply ingrained in human nature. Extensive evidence demonstrates that humans possess a remarkable capacity to prioritize fairness over personal gain \cite{civai2010irrational,sanfey2003neural}. In various decision-making scenarios, individuals have consistently exhibited a willingness to sacrifice their own interests to uphold a sense of fairness within their environment \cite{koenigs2007irrational}. They may work to mitigate perceived inequality, even going as far as foregoing personal benefits so that others may benefit~\cite{fehr1999theory}. 

With robots being increasingly placed in situations where their decisions could disproportionately affect people, there is increased pressure to ensure that they behave fairly. Scenarios such as caregiving robots disproportionately distributing their support among patients \cite{yew2021trust} and factory robots scheduling tasks unevenly across human workers \cite{ranz2017capability} demonstrate that robots can promote differential treatment between people and trigger fairness judgments, which ultimately affects the intimacy \cite{claure2023social} and the trustworthiness of robots \cite{de2007effects}. 


Despite the growth of interest in fairness research in Human-Robot Interaction (HRI), the way fairness is currently studied is limited in two ways. First, prior research only measures fairness perceptions about a robot at the conclusion of an interaction \cite{claure2020multi, fraune2019human}, as shown in Figure~\ref{fig:overview}(a). Second, there is no framework that guides the study of the underlying factors that may lead to a robot being perceived as unfair. 

To address the above gaps, we examine how fairness perceptions vary \textit{over time} during multi-party human-robot interactions, as in Figure~\ref{fig:overview}(b). Additionally, we explore using the \textit{Fairness Theory} by \citet{folger2001fairness} in HRI. In the context of Organizational Justice, the theory proposed that human fairness perceptions depend on three interconnected components: reduced welfare, conduct, and moral transgression. We apply the theory by investigating if the three components help predict how humans perceive the fairness of robot behavior. 

More specifically, we conducted a study to investigate how fairness judgments update over the course of an interaction when a robot unfairly supports a human or a competitor robot. Participants competed in a multi-player Space Invaders game against a robot while a second robot, Nao, provided support to the players, potentially in unequal amounts. Through this study, we sought to understand how the timing of the unfair support and the beneficiary of the unfair support influence how fairness perceptions change over time. Further, we examined using the Fairness Theory to better understand  the underlying factors that drove fairness perceptions towards the Nao robot in our study. 


In summary, our work makes three primary contributions. First, to our knowledge, we are the first to explore how perceptions of fairness can change over the course of a situated interaction with a robot. We study this phenomenon using video self-annotations \cite{zhang2023self}. Second,  we show how fairness perceptions towards a robot are impacted by the timing of unfair support. Further, we demonstrate how a robot's unfair support shapes group perceptions and impacts trust toward the robot. 
Lastly, we explore using the three components of the Fairness Theory \cite{folger2001fairness} to predict when humans consider robot actions to be unfair. Our findings pave the path to exciting, future research directions on momentary fairness perceptions.



%


\section{Related Work}
\textbf{Fairness in Artificial Intelligence (AI).} 
A significant amount of work in algorithmic fairness studies how to make Artificial Intelligence (AI) systems less prone to biases that may negatively shape their decisions (e.g., ~\cite{barocas2017fairness,mitchell2021algorithmic,halevy2021mitigating,kim2021age}). 
For example, 
in hiring practices, AI decisions have discriminated against female candidates \cite{datta2014automated,dastin2018amazon}. Individuals of color have shown to not be recognized by certain computer vision algorithms \cite{buolamwini2018gender}. The causes of these unfair outcomes often lie in human bias embedded across different parts of the algorithmic development pipeline. For instance, human prejudice can be embedded into training datasets 
\cite{barocas2017fairness}. 


A fair AI algorithm circumvents human biases and avoids utilizing attributes related to protected groups like gender or race \cite{pessach2020algorithmic}. This means that it neither amplifies inequalities nor imposes disproportionate negative impacts on specific groups. Mathematical fairness objectives aim to formalize this idea. For instance, 
disparate impact requires that positive predictions are identical across protected and non-protected groups~\cite{barocas2016big}. Demographic parity requires that a positive prediction is assigned to groups at a similar rate~\cite{jiang2021generalized}. 

Other research has explored behavioral responses to AI decisions. 
People tend to respond negatively when an AI system displays behaviors that disproportionately support an individual \cite{saxena2019fairness}, allocates resources unequally \cite{ClaureReinforcementTeams,claure2023social}, or  promotes differential treatment \cite{uhde2020fairness}. HRI research, including this work, often builds on these prior efforts when investigating fairness perceptions toward robots, as discussed next.%

\vspace{1em}
\noindent
\textbf{Fairness in HRI.}  Human responses to perceived unfairness in robots have shown to be intense and can be emotionally charged \cite{arnold2018observing,litoiu2015evidence,short2010no, chang2020defining,chang2021unfair}. Broadly speaking, this response is driven by the context of the unfair situation and how the robot behaves \cite{arnold2018observing}. For instance, an encounter with a cheating robot led individuals to calibrate their engagement levels and examine whether such conduct was a malfunction or a calculated decision \cite{short2010no, litoiu2015evidence}. Moreover, the consequences of a robot behaving unfairly extend to the social dynamics that enable individuals to collaborate with one another. Previous studies suggest that people tend to perceive less closeness \cite{claure2023social} and trust \cite{claure2020multi} toward a robot when they receive less support from it. Typically, these works on trust tend to focus on the performance dimension of trust. While this focus on performance is critical, it is insufficient for studying trust in human-robot interactions. Trust is a multi-dimensional concept, encompassing both performance and moral aspects \cite{Malle2013}. The inherent human biases that can appear in interactions with robots raise concerns about the moral dimension of trust in robots \cite{howard2018ugly,ogunyale2018does,bartneck2018robots,claure2024designing, nashed2023fairness}.

However, the leap from algorithms to robotics in fair decision-making research introduces layers of complexity that warrant distinct consideration. Robots, unlike static algorithms, possess the capability to operate in both private and public spheres \cite{kaminski2014robots,kanda2005communication,kanda2010communication}, engaging directly with individuals and groups \cite{sebo2020robots}, which can lead to evolving tasks over time. This dynamic nature of robotics introduces novel fairness challenges, such as the risk of physical harm stemming from biased algorithms—for instance, a self-driving car's failure to detect darker-skinned individuals \cite{wilson2019predictive}. Furthermore, the context-sensitive aspect of robotic tasks affects the perception and implementation of fairness. In certain scenarios, an equal distribution of resources by robots may be desired \cite{hamann2014meritocratic}, whereas in other contexts, a strategy that considers individual contributions and needs could be more equitable \cite{moore2009fairness}. This stands in contrast to AI algorithms used in sectors like finance or criminal justice, where decisions are often one-time occurrences. The continuous decision-making processes that take place in robotics, especially in social settings, coupled with the effects of robot's physical embodiment on users, necessitate a more intricate and context-aware approach to fairness \cite{howard2020robots}. 

Although research on fairness perceptions in HRI is growing, the current body of work remains limited to measuring fairness perceptions about a robot at the conclusion of an interaction. For example, fairness is commonly considered a static attribute of a robot \cite{litoiu2015evidence,claure2020multi, fraune2019human,chang2021unfair,arnold2018observing}. Meanwhile, research in organizational psychology suggests that human perceptions of fairness can update over time as people's understanding and experiences accumulate over time  \cite{jones2013perceptions,holtz2009fair}. Drawing further inspiration from work on modeling trust on a moment-by-moment basis \cite{chen2018planning,guo2021modeling,yang2021toward,chi2023people}, we sought to investigate questions around the evolution of fairness perceptions. 
Thus, our work is driven by the research questions (RQs): 


\vspace{0.5em}

\noindent\textit{\textbf{RQ1:} Do humans’ fairness perceptions of a robot’s behavior vary (across time) during their interaction?}

\vspace{0.5em}

\noindent\textit{\textbf{RQ2:} How does the timing of a robot’s unfair treatment impact humans’ perception of the overall interaction?}
\vspace{0.5em}

We studied the above questions on a multi-party interaction involving competitive gameplay in Space Invaders, given that this environment has been useful to study human-agent interactions in the past \cite{large2020studying,candon2022perceptions} and can be easily adapted to HRI \cite{candon2023verbally}. This competitive setting also motivated us to ask:

\vspace{0.5em}
\noindent \textit{\textbf{RQ3:} Does the recipient of the unfair behavior from a robot (self vs. competitor) influence human fairness perceptions?}

For the latter question, we focused our work on investigating situations where the competitor was another robot, which made the interaction experience more consistent across study sessions in comparison to situations where the competitor was another human. Studying situations where the competitor is a human is interesting future work.

\vspace{1em}
\noindent
\textbf{Fair Robot Behavior Generation.}
 The importance that humans place on fairness within collaborative settings \cite{colquitt2015justice} has led to a recent line of work that is focused on enabling robots to be able to make fair decisions \cite{claure2020multi,chang2020defining,chang2021unfair,short2010no,brandao2020fair}. This is often studied in HRI from a distributional perspective, where a robot has to decide how to allocate different forms of resources (gaze \cite{mutlu2009footing}, attention \cite{tennent2019micbot}, artifacts \cite{jung2020robot}) across interactants. The focus on fair resource allocation stems from the fact that humans are sensitive to how much they receive compared to others \cite{adams1965inequity} and behave according to how fairly they feel treated \cite{bettencourt1997contact}.

Prior HRI work has proposed specific solutions for fair resource allocation within different contextual settings \cite{claure2020multi,brandao2020fair,chang2020defining}. For instance, Claure et. al \cite{ClaureReinforcementTeams} proposed a multi-armed bandit algorithm for a robot  to support team members with different levels of capabilities  while considering fairness. 
Others have explored methods for the fair distribution of attention and support by robots during navigation missions \cite{brandao2020fair,zhounoise,brandao2021socially}. 

Despite the above progress, there is a lack of a general framework for modeling fairness perceptions in HRI, which could inform the future development of more general algorithms for fair robot behavior generation. To address this gap, we explore applying a Fairness Theory \cite{folger2001fairness} from Organizational Justice to HRI, as described in the next section.



   


\vspace{1em}
\noindent
\textbf{Fairness Theory.}
The theory by \citet{folger2001fairness} proposes that there are three interconnected components holding someone or something accountable: \textit{reduced welfare}, \textit{conduct}, and \textit{moral transgression}. 
%
For people to feel that an event is unfair, there must be an unfavorable event or condition that should threaten an individual's well-being, be it material or psychological (\textit{reduced welfare}). Once the responsible party behind the unfavorable event is identified, an evaluation of their potential to behave differently takes place (\textit{conduct}). Finally, an assessment is conducted to determine whether this harmful injury violates the ethical principle of interpersonal treatment (\textit{moral transgression}). These interconnected steps collectively help individuals assess whether or not the transgression that occurred is fair.

While the Fairness Theory has been applied 
to study human responses to unfairness in the workplace \cite{keashly2010faculty,  nicklin2011importance}, its extension to HRI seems timely, especially as robots become more prominent in the workplace. Specifically, we suspected that the theory could provide robots a pathway to predict fairness perceptions toward them. 
Because the theory deconstructs fairness judgments into three distinct yet measurable components, we ask:

\vspace{0.5em}
\noindent \textit{\textbf{RQ4:} Can perceptions for the three components of fairness (reduced welfare, conduct, and moral transgression per the Fairness Theory \cite{folger2001fairness}) be used to predict momentary fairness perceptions towards a robot?} 

\section{Method}



Driven by the research questions outlined in the prior section, we conducted a user study in which  participants played several rounds of a multi-player Space Invaders game with two robots. One robot, Shutter, served as a competitor against the participant. The other robot, Nao, played in a supporting role. Nao could help either the participant or Shutter during the game. Our work mainly focused on understanding how participants perceived the behavior of the Nao robot, which sometimes distributed its help in unfair ways (in unequal amounts) in the game.

The study had a mixed $2 \times 2$ design. One factor pertained to the \textit{timing of unfair actions} by a robot (either in the early or late stages of a game) and the other related to the \textit{beneficiary} of such actions (either the participant in the study or a competitor in the game). The timing of unfair actions served as a within-subjects variable, ensuring each participant experienced both timing conditions. The beneficiary was treated as a between-subjects variable, exposing each participant to only one of the two conditions. 
In relation to our research questions, we hypothesized that: 

\vspace{0.5em}

\noindent \textbf{H1}: Unequal distribution of help by the Nao robot will result in lower momentary perceptions of fairness in comparison to a more equal distribution of help during the Space Invaders game.


\vspace{0.5em}
\noindent \textbf{H2}: Unequal distribution of help by the Nao robot later in a game will have a more detrimental impact on the overall fairness perceptions that participants report for the robot for a whole interaction than unequal distribution earlier in the game.

\vspace{0.5em}
\noindent \textbf{H3}: When the beneficiary of the robot's unfair behavior is a competitor, humans will perceive the Nao robot as less fair in comparison to when they are the beneficiary and receive more help.



\vspace{0.5em}
\noindent \textbf{H4}: Momentary perceptions of unfair robot behavior can be predicted based on perceptions of the three components of fairness (reduced welfare, conduct, and moral transgression) from \cite{folger2001fairness}.

\vspace{0.5em}
The first hypothesis aimed to address RQ1, as we expected fairness perceptions to vary as a function of Nao's behavior in the game. This is motivated by work in psychology which points to how variability in treatment can lead to changes in fairness judgments \cite{matta2017consistently,jones2013perceptions}. The second hypothesis (in relation to RQ2) was motivated by humans' recency effect, a cognitive bias in which items, ideas, or arguments that are experienced last are remembered more clearly than those that came first \cite{TURVEY2012495}. This effect has been observed in HRI in the past, e.g., in the study of trust in automation \cite{yang2017evaluating}. The third hypothesis stated that the recipient of the unfair behavior from a robot would influence fairness perceptions (RQ3), given that people have shown to be sensitive to differential treatment \cite{leventhal1976distribution,dulebohn1998employee,bartol2002encouraging}. The last hypothesis (in relation to RQ4) was motivated by prior use of the Fairness Theory \cite{folger2001fairness} to analyze human-human interactions. Validating these hypotheses is important as robots continue to permeate environments where they have tangible impact on humans, necessitating an understanding of when their actions may be perceived as unfair \cite{brandao2020fair}. 







\subsection{Materials}

\begin{figure*}[t]
    \centering
    \includegraphics[width=1.0\linewidth]{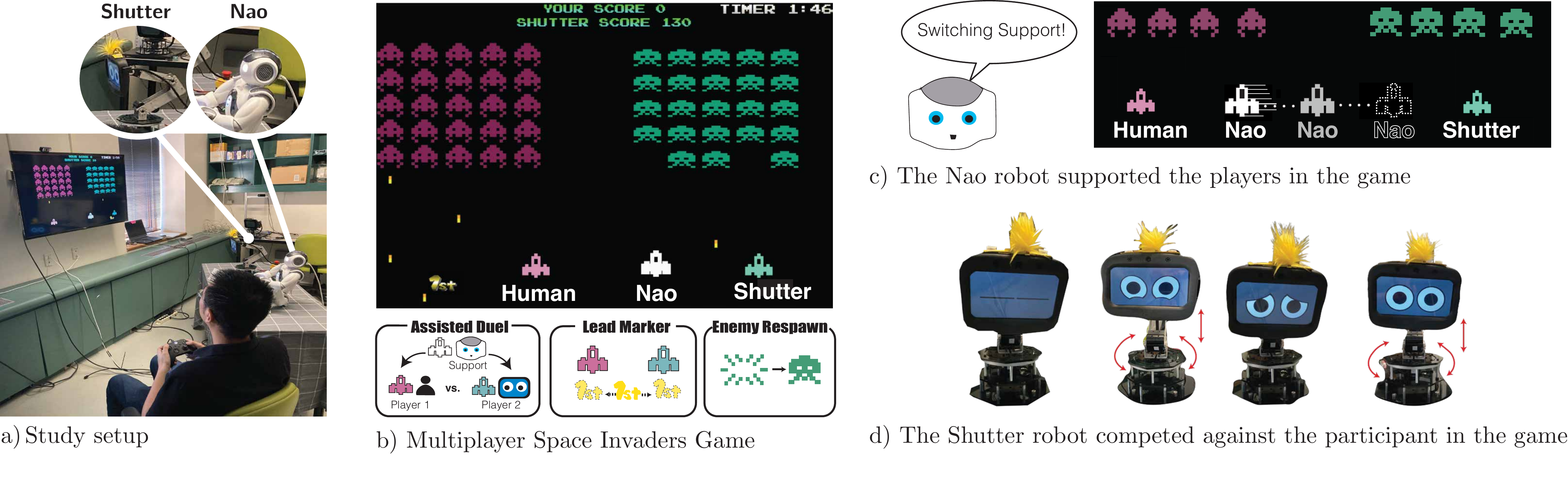}
    \caption{The study took place in a laboratory (a). The participant played a multi-player Space Invaders game game with two robots (b). Nao's spaceship (white) supported players by eliminating enemies on their side of the game screen. The lead marker moved to the side of the player with the highest score. Enemies in the game  reappeared after being eliminated. The Nao robot moved its head to face the player whose spaceship it supported and made  comments to convey that it was trying to help them (c). The Shutter robot was expressive through four different behaviors controlled via a behavior tree control architecture (d). }
    \label{Game_and_robot}
\end{figure*}

We employed two different robots in our study (as shown in Fig.~\ref{Game_and_robot}) to reduce perceptions of them being part of a team because of their embodiment \cite{fraune2020our}. One robot was a Nao robot by Aldebaran, which looks highly humanoid and can communicate verbally and gesture. The other robot, Shutter, was  also designed for social interactions \cite{adamson2020designing,lew2023shutter} but had less of a human form. Shutter is an open software and hardware platform  with a 4 degrees of freedom arm and a facial display \cite{thompson2024shutter}. It leverages behavior trees \cite{colledanchise2018behavior} for high-level robot control.

We adapted the Space Invaders platform from \citet{candon2022perceptions} to support multi-player interactions. We chose this platform for three reasons. First, this platform could be used to recreate the commonly observed scenario in HRI where a robot makes decisions on how to distribute resources \cite{claure2020multi,jung2020robot,shah2011improved}. Second, different versions of this platform have been successfully used in prior research \cite{large2020studying,candon2022perceptions,candon2023verbally}, helping us build and expand on prior work that studies how humans perceive the helping behavior of an agent -- the Nao robot in our case. Finally, Space Invaders has a rich history in various fields including psychology \cite{mateas2003expressive} and artificial intelligence \cite{ho2020achieving}. 

The multi-player Space Invaders game developed for this study employs three distinct spaceships controlled by three players: the Shutter robot, the Nao robot, and the human participant. Each player commands an individual spaceship, differentiated by color -- red for the human player, white for Nao, and green for Shutter (as in Figures \ref{Game_and_robot}a and \ref{Game_and_robot}b). The adversaries in the game are represented as alien spaceships and organized into two distinct clusters on the display. This configuration allowed for a concentrated focus by the Shutter robot on the adversaries on the right side of the screen and by the human player on the adversaries on the left. The alien adversaries were programmed to reappear after being eliminated, so enemies could be destroyed continuously until the game ended at the two-minute mark. 

The participant and Shutter competed in the game to eliminate as many alien spaceships as possible within their respective sides of the screen. 
We chose to use Shutter as a competitor for the participant because preliminary pilot tests highlighted challenges in maintaining consistency across sessions with a human competitor. Incorporating Shutter enabled tighter control over experimental variables, ensuring that any change in fairness perceptions could be more confidently attributed to the actions of the Nao robot. 

The Nao robot served as a support player, able to aid both Shutter and the participant by eliminating enemies in their respective section of the game screen. Adversaries eliminated by Nao contributed to the score of the player whose designated screen section they occupied. Our main hypotheses were then focused on how humans perceived the behavior of the Nao robot.

To emphasize the game's competitive nature, a lead marker moved toward the player with the highest score (Fig.~\ref{Game_and_robot}b). Additionally, the shooting speed of the robots' spaceships was imperceptibly faster than the shooting speed of the human player. This made the experience   similar for participants with different skills in the game. Also, it helped ensure that when Nao focused on helping a player, the beneficiary had a major advantage in the game.

The Nao and Shutter robots were expressive during the Space Invaders game. Nao was programmed to turn its head toward the player that it supported and randomly uttered certain phrases affirming its behavior. For example, Figure~\ref{Game_and_robot}c illustrates how Nao expressed  switching support toward a player. Nao would also highlight the winner of game rounds, as shown on the game screen, at the conclusion of a game. Shutter 
showed different behaviors depending on the state of the game, such as a sad emotion when it lost a game round (as in Figure~\ref{Game_and_robot}d).

\subsection{Procedure}
\begin{figure*}[t]
    \centering
    \includegraphics[width=1.0\linewidth]{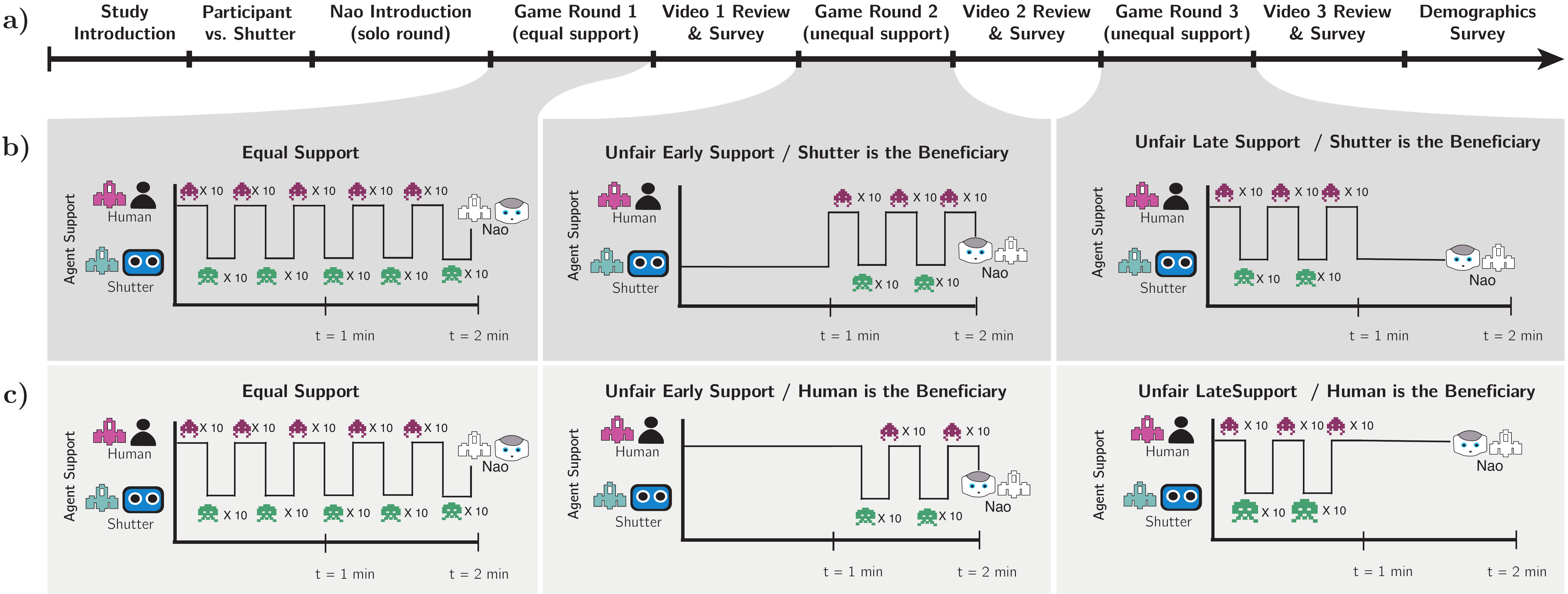}
    \caption{Timeline of a study session (a). The study began with an introduction to the robots and two practice games to learn about Space Invaders. Then, during Game Round 1, participants experienced a 2 minute game where Nao equally supported each player by switching its support when it eliminated 10 enemies on one side. In game rounds 2 and 3, the participants experienced Nao providing unequal support. For the first or the second half of these games, Nao focused on helping only Shutter (b) or the human participant (c). We measured  impressions of the experience for each game round. Early and late unfair support (in the second and third game rounds) was counterbalanced across participants to reduce potential ordering effects. }
    \label{Study_Design}
\end{figure*}

Figure \ref{Study_Design} details the sequence of events that took place in the study. After signing a consent form, the participants were introduced to the robots and were informed of the rules of the Space Invaders game. This information was provided by the experimenter and through a webpage that was shown on a television screen placed in front of the players (as in Fig.~\ref{Game_and_robot}a). The participants were also informed that 
they would receive a bonus of US\$1 for every game round in which they got a higher score than Shutter in the game. 

The participants then learned about the robots' capabilities. First, they took part in a practice game where they played for 15 seconds only with Shutter. Afterward, they watched Nao play a 15 second game by itself, demonstrating how it could support either player by destroying enemies on both sides of the game screen. 

After the introductory explanations, the experimenter left the room and the participants completed the first game round with Nao and Shutter which lasted 2 minutes. In this game, the participants observed Nao switching its support equally between both players after eliminating  ten enemies on a given side of the game screen.  Upon the completion of the game, the experimenter returned to the room and asked the participant to review a video of their last game on a nearby computer and answer a few survey questions about their experience, similar to the self-annotation procedure from \citet{zhang2023self}. The participants watched the video of their gameplay, which stopped at the 5, 60, and 119 second mark. At these critical points in time, the participants answered survey questions about their perceptions of Nao and their closeness to the robots. 

The above procedure was repeated two more times: the participant played another game with the robots and then reviewed the game video, answering questions about their experience. 

Upon completing all three game rounds and corresponding video surveys, participants were given a final survey with questions about their overall experience. The participants were compensated US\$10 dollars for completing the study activities and given an additional bonus for every game round they won. Thus, they received between \$10 - \$13 in total for participating in the study.

\vspace{0.5em}
\noindent
\textbf{Timing of Unfair Actions}: During game rounds 2 and 3, the participants observed the Nao robot distributing its help unequally toward the players. In one game, Nao acted unfair \textit{early} in the round and in the other, it acted unfair \textit{late} in the round. In the case of early-round unfairness, Nao allocated its support exclusively to one player for the first half of the round (1 minute). For the remaining duration of the round, Nao switched its support equally between both players. For late-round unfairness, Nao distributed equal support to both players for the initial half of the round (1 minute) and, for the latter part of the round, Nao solely supported one player. The order of 
the early and late unfair actions was counterbalanced to reduce potential ordering effects. 

\vspace{0.5em}
\noindent
\textbf{Beneficiary of Unfair Action}: Participants were assigned to observe either themselves or the Shutter robot benefiting from Nao's gameplay actions during game rounds 2 and 3 (as illustrated in Figures~\ref{Game_and_robot}a and \ref{Game_and_robot}b). This assignment was semi-random to balance this between-subjects variable. For example, if a participant was chosen to benefit from Nao's unfair actions, they were the sole recipient of Nao's support when it distributed support unequally. 

\subsection{Measures}

\begin{description}[leftmargin=*,itemsep=0.5em]

\item[Fairness Perceptions:] We used survey items to measure \textit {momentary fairness} (during a game) and  \textit{overall fairness} (after the game ended). Momentary fairness corresponded to human perceptions of fairness at a certain point during an interaction. We measured it by asking the participants to rate how much they agreed with the statement: ``Nao's support towards me was fair'' on a 5-point Likert responding format (1 being ``strongly disagree'' and 5 being  ``strongly agree''). This question was presented to participants at the beginning, middle, and end points of a game video. The question specifically asked participants to consider the time up to the point where the video stopped to answer the question in the corresponding video survey. 

\setlength\parindent{12pt}Overall fairness corresponded to participant's perception of the fairness of the Nao robot considering their overall experience in a given game of Space Invaders. To capture it, we asked the participants to consider an entire game round and to indicate how much they agreed with the statement: ``Nao's support towards me was fair'' on a 5-point Likert responding format (1 being lowest as before). This survey item was presented to the participants at the conclusion of each video review activity. 

\setlength\parindent{12pt}In order to capture how people were conceptualizing Nao's decision on who to support, we asked the open-ended question "What factors do you think Nao considered for making its decision on who to support?" after all the game rounds had concluded. 
\setlength\parindent{12pt} To analyze responses, we first developed a coding scheme consisting of nine distinct categories. These distinct categories emerged from a thorough examination of the responses. Then, two reviewers took part in a three step process to analyze the results. (1) First, both reviewers engaged in a preliminary coding exercise that encompassed 10\% of the open-ended survey responses. This initial phase served both as a calibration exercise and as a means to foster a shared understanding of the coding framework. Then, they met to assess inter-rater alignment and to resolve any discrepancies in the interpretation or application of the coding scheme. (2) We  expanded the task to include 25\% of the survey responses, thereby allowing the reviewers to apply the coding scheme to a broader and more representative sample of the data. Another session was held afterward to ensure consistency and to refine the coding scheme where necessary. (3) In the final phase of analysis, our reviewers independently coded the entirety of the open-ended responses. This analysis culminated in a final consensus meeting resulting in a high level of inter-rater reliability (with a Cohen’s Kappa coefficient of 0.86). If coders assigned different categories to a response, we considered all the annotations as valid labels. For all the steps of the coding process, reviewers were told that they were allowed to select multiple categories for each open-ended response.
\item[Perceived Closeness:] As an exploratory measure, we investigated the social consequences of Nao's unfair actions. In particular, we used the Inclusion of Others in the Self (IOS) scale \cite{aron1992inclusion} to measure the perceived closeness between the  participants and Nao, and between Nao and Shutter. For the former perceptions, the IOS scale was presented with seven images of increasingly overlapping circles labeled ``you’’ and ``Nao’’. The participants rated which images best represented their relationship on a Likert responding format (from 1 being very distant with no circle overlap  to 7 being very close with a high degree of overlap between the circles). For the latter perceptions, the circle labeled ``you'' was instead labeled ``Shutter'', so that the participants rated the relationship between the robots (as depicted in the supplementary material). The IOS questions were administered after each game. Importantly, while our research was not specifically focused on closeness and we did not have specific research questions in relation to these measures, we generally expected the participants to feel that Nao was closer to the player that benefited from the unfair support. This expectation was motivated by prior research that shows that robots can influence the behaviors and attitudes exchanged between two other interactants in HRI \cite{gillet2024interaction}. 


\item[Trust:] Similar to perceived closeness, our research was not specifically motivated by research questions about trust, but we measured trust perceptions toward the Nao robot because the relationship between fairness and trust is not yet fully understood in HRI. Using the Multi-Dimensional Measure of Trust (MDMT) \cite{mdmtv2}, we measured two distinct dimensions of trust toward Nao: Performance Trust and Moral Trust. A ``Does Not Apply'' (NA) option was included for each question as in the original MDMT survey, allowing the participants to indicate if they thought the rating attribute did not apply to Nao. Internal consistency for each scale, excluding the NA responses, was satisfactory (performance trust resulted in Cronbach’s $\alpha$ = 0.92; moral trust led to Cronbach’s $\alpha$ = 0.90). Assuming that participants considered an equal distribution of help as optimal and fair for the Nao robot, we expected unfair support by the robot to affect human perceptions of moral trust toward it. Also, we expected the behavior of Nao in the equal support game to result in higher performance trust perceptions than in the other games.

\item[Fairness Components:] 
We asked distinct questions to capture the three components of fairness proposed by  \citet{folger2001fairness}: \textit{reduced welfare, conduct, and moral transgression}. These questions were targeted to gauge participants' perception of fairness toward Nao's actions in relation to their experience in the game. To measure perceptions of reduced welfare, we asked the participants to rate how much they agreed with: ``Nao's gameplay strategy significantly reduced my chances of success" on a 5-point Likert response format (1 being lowest). To measure if the participants believed that Nao could have taken a different strategy (conduct), we asked them: ``How strongly do you feel that the robot had feasible alternatives to its actions in this part of the game?''  on a 5-point Likert responding format with 1 being ``Not at All'' and 5 being ``Very Strongly''. Finally, to measure moral transgression, we asked the participants to think about the social appropriateness of Nao's actions in the competitive context. In particular, we asked: ``How would you rate the social appropriateness of the robot's decisions in the context of the competition between yourself and Shutter?'' on a 5-point Likert responding format (1 being lowest). 
These three questions for reduced welfare, conduct, and moral transgression were presented to participants at the beginning, middle, and end points of a game round video while they reviewed their experience.


\end{description}

\subsection{Study Participants}
We recruited 40 participants through flyers and word of mouth. Of the 40 people, 20 of them experienced unfair support by Nao during gameplay while 20 experienced Shutter benefiting from unfair support by Nao. The order of the early and late unfair actions was counterbalanced. The demographic spread included 19 participants who identified as female, 19 as male, 1 as nonbinary, and 1 preferred not to disclose their gender. Additionally, 27 of the participants had experience with Space Invaders, 9 had no experience with the game, and 4 were unsure. Our local Institutional Review Board approved the study and all participants signed a consent form agreeing to participate. A breakdown of the participants can be seen in Table 1.

\begin{table}[h!p]
\centering
\caption{Participant Demographics by Condition.}
{\footnotesize
\begin{tabular}{lccccccc}
\toprule
Beneficiary & Order & N & Male & Female & Other& Age ($\mu \mp \sigma$) \\
\midrule
Human & Early,Late & 10  & 7 & 3 & 0 & 26.7 $\mp$ 4.64\\
Human & Late,Early & 10  & 4 & 5 & 1 & 24.3 $\mp$ 5.10\\
Shutter & Early,Late & 11 & 6 & 5 & 0 & 30.2 $\mp$ 15.1 \\
Shutter & Late,Early & 9 & 2 & 6 & 1 & 24.8 $\mp$ 3.96\\
\midrule
 & All & 40 & 19 & 19 & 2\\
\bottomrule

\end{tabular}
}
\label{tab:regression}
\end{table}

\subsection{Manipulation Check}

We confirmed that the participants were aware of the level of support they received from Nao through a survey question after each game round: 
``Reflecting on the recently completed round, please rate your perception of the relative support that Nao provided to you and Shutter. Choose the statement that best aligns with your experience.'' Participants then responded on a 5-point Likert responding format, with 1 being ``Nao provided significantly less support to me than Shutter'' and 5 being ``Nao provided significantly more support to me than Shutter.'' We used a mixed linear model estimated with Restricted Maximum Likelihood (REML) \cite{corbeil1976restricted} to analyze these ratings. The model considered Beneficiary of Unfair Support (Shutter or the Participant),  Game (Equal Support, Early Unfairness or Late Unfairness), and Order (Early Unfairness First or Late Unfairness First) as main effects and participant ID as a random effect. 

We found a significant difference in the relative support ratings by Beneficiary of Unfair Support, $F[1,36]= 359.68, p<.0001$. 
A Student's t-test confirmed that participants who benefited from unfair support correctly perceived that Nao supported them more than Shutter $(M=4.15, SE=.15)$ compared to participants who received less support than Shutter $(M=1.81, SE=.15)$. The interaction between Beneficiary of Unfair Support and Game also had a significant difference on the relative support ratings, $F[2,74]= 150.69, p<.0001$. A Tukey Honestly Significant Difference (HSD) post-hoc test on the interaction between the Beneficiary and Game showed three significantly different tiers for relative support ratings. The ratings were highest for the two games in which the participants received unfair support (Late Unfairness: $M=4.91, SE=.06$ ; Early Unfairness: $M=4.82, SE=.11$), followed by the two games of Equal Support $(\text{Human: } M=2.73, SE= .19; \text{ Shutter: } M=3.11, SE= .24)$. The lowest ratings were for the two games in which Shutter received unfair support $(\text{Late Unfairness: } M=1.17, SE= .09; \text{Early Unfairness: } M=1.17, SE= .09)$.
Finally, we also found a significant difference on relative support ratings by the interaction between Game and Order, $F[2,74]= 3.26, p=.044$, though a Tukey HSD test showed no significant differences. 
The first two results indicate that our manipulations worked as intended.  

\section{Results}
This section presents our study results. We used linear models estimated with Restricted Maximum Likelihood (REML) \cite{corbeil1976restricted} to  examine participant's responses to video reviews and surveys. In these models, the Beneficiary of Unfair Support (Shutter or the Participant) was considered as a main effect. When the measures were repeated by Game (Equal Support, Early Unfairness, or Late Unfairness) and Round Period (Beginning, Middle, End), we included these variables as main effects as well as the Order in which the participants experienced the timing of unfairness  (Early Unfairness First or Late Unfairness First). Additionally, we added participant ID as a random effect to the model. We conducted post-hoc Student's t-tests and Tukey HSD tests when appropriate.

\subsection{Momentary Fairness Perceptions}
\label{ssec:results_momentary_fairness}

The analysis on momentary fairness perceptions  showed that Game (\textit{F}[2,304] = 26.05, \textit{p}~$< .0001$) and Round Period (\textit{F}[2,304] = 5.25, p~$= .005$) had significant effects on the ratings. 
As expected, the momentary fairness ratings were significantly higher for Equal Support $(M=3.35, SE=.12)$ than for Early Unfairness $(M=2.32, SE=.12)$ and Late Unfairness $(M=2.62, SE=.12)$. That is, the participants perceived the behavior of the Nao robot as generally more fair in the Equal Support game than in the other games they experienced. 
Additionally, we found that momentary fairness ratings at the end of a game round were significantly lower $(M=2.49, SE=.11)$ than at the beginning $(M=2.94, SE=.11)$ or middle $(M=2.86, SE=.11)$ of a game round. This finding aligned with our idea that fairness perceptions about a robot's behavior can change over time. 

\begin{figure}[t!p]
    \centering
    \includegraphics[width=.88\linewidth]{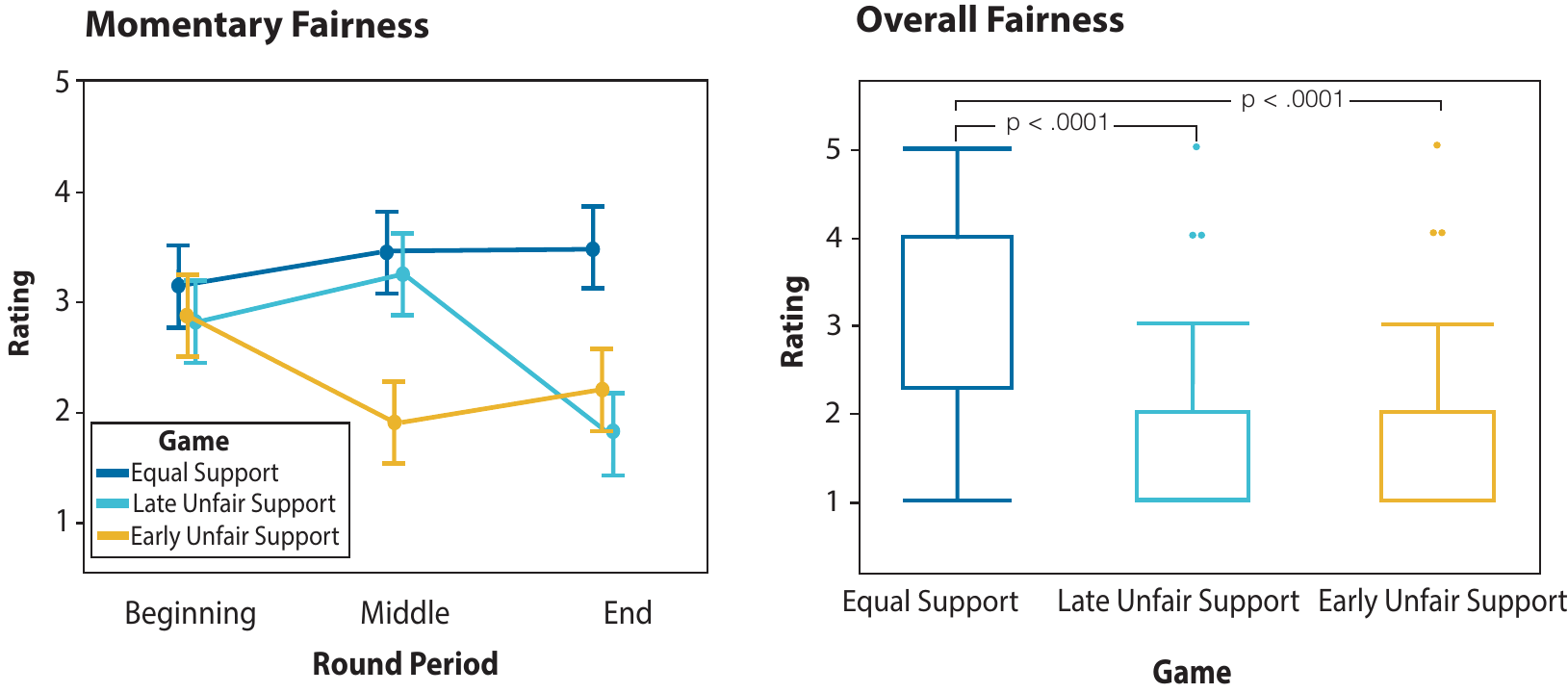}
    \caption{ Left: Average momentary fairness ratings (error bars are standard error). Right: Overall fairness ratings. The boxes represent the interquartile ranges of ratings. The dots outside the boxes denote outliers.}
    \label{fig:momentary_Fairness}
\vspace{-1em}
\end{figure}

Importantly, there was also a significant interaction effect between Round Period and Game on the momentary fairness ratings, \textit{F}[4,304] = 10.43, \textit{p}~$< .0001$ (as illustrated in Figure ~\ref{fig:momentary_Fairness}). 
The post-hoc test showed no significant differences among the ratings for the Equal Support game. 
However, when the unfair support came early in the game round, the ratings for the beginning of the game (\textit{M} = 2.88, \textit{SE} = .189) were significantly higher than for the middle (\textit{M} = 1.92, \textit{SE} = .189, \textit{p} = .005) of the game (after Nao had helped one player more than the other). The ratings at the end of the game (\textit{M} = 2.22, \textit{SE} = .189) were not significantly different from those at the beginning or the middle. 
When the unfair support came late in the game, the ratings at the beginning  and middle of the game were not significantly different. But the ratings at the end of the game (\textit{M} = 1.82, \textit{SE} = .189) were significantly lower than at the beginning (\textit{M} = 2.82, \textit{SE} = .189) 
and the middle (\textit{M} = 3.28, \textit{SE} = .189) 
of the game. 
This last result partially supports our first hypothesis (H1), where we expected unequal distribution of help by Nao to result in lower momentary fairness perceptions in comparison to a more equal distribution.

In relation to our third hypothesis (H3), we expected perceptions of momentary fairness to be impacted by whether the participants or Shutter benefited from Nao's unfair actions. However, our analysis on momentary fairness perceptions did not find a significant main effect for 
the Beneficiary of Unfair Support (p~$ = 0.31$). 

\subsection{Overall Fairness Perceptions}

The analysis on overall fairness showed that Game (\textit{F}[2,74] = 69.5,\textit{p} < .0001) had a significant effect on ratings of fairness considering entire game rounds. 
A Tukey HSD revealed that participants rated Nao's actions significantly more fair in the Equal Support game $(M=3.48, SE=.18)$ than in both the Late Unfairness $(M=1.63, SE=.16)$ and Early Unfairness $(M=1.60,SE=.16)$ games. This result did not support our second hypothesis (H2), in which we expected an Unequal distribution of help by the Nao robot later in a game to have a more detrimental
impact on the overall fairness perceptions than an unequal distribution earlier in the game.

The analysis on overall fairness ratings showed a trend for Beneficiary of Unfair Support having an effect on the results $(p=.07)$. Participants who did not receive the unfair support (i.e., those that experienced Shutter being the Beneficiary of Unfair Support) rated Nao's behavior as less fair $(M=1.96, SE=.18)$ than participants who were the Beneficiary of the Unfair Support $(M=2.45, SE=.17)$. This trend was in line with our third hypothesis (H3).

The open-ended responses revealed differences in how participants conceptualized Nao's decision-making when it disproportionately supported them over Shutter. Table 2 describes the categories that emerged from our qualitative analysis of the participant's responses. Participants were more likely to attribute calculated strategy and bias to Nao's decision-making when the robot Shutter was the beneficiary of support. Conversely, disproportionate support for humans was more often seen as dynamic and potentially justified within the context of the game. When Nao disproportionately supported Shutter, participants felt that Nao was acting strategically (19.1\%) but indicated a stronger sensitivity or concern toward unfair advantage given to their robot competitor (23.4\%). For example, P777 commented, \textit{``Nao favored the robot every single interaction -- I think Nao was programmed to assist the robot every time and did not pay attention to the score. Perhaps time was one factor (...).''} Others stated that Nao's unfair support of Shutter was strategic and, as P491 and P297 stated, time-based. When Nao unfairly supported the participants, they justified its behavior by noting that Nao was acting strategically (16.3\%) and even adjusting its strategy by round (16.3\%). Several participants mentioned how they believed Nao's support was strategically decided based on factors related to game performance. For instance, P101 suggested that \textit{``Nao considered who was winning and the overall score (...).''} P505 commented how Nao supported \textit{``(...) the player that had fewer enemies remaining on the field at any time.''} These results provide additional support for H3.

\setcounter{table}{1} 
\begin{table}[t!p]
    \centering
    \caption{Qualitative codes and percent of participants' responses that mentioned the code. The last two columns indicate the distribution of responses relative to the total number of responses in that condition.}
    \label{tab:qualitative_codes}
    \includegraphics[width=.95\linewidth]{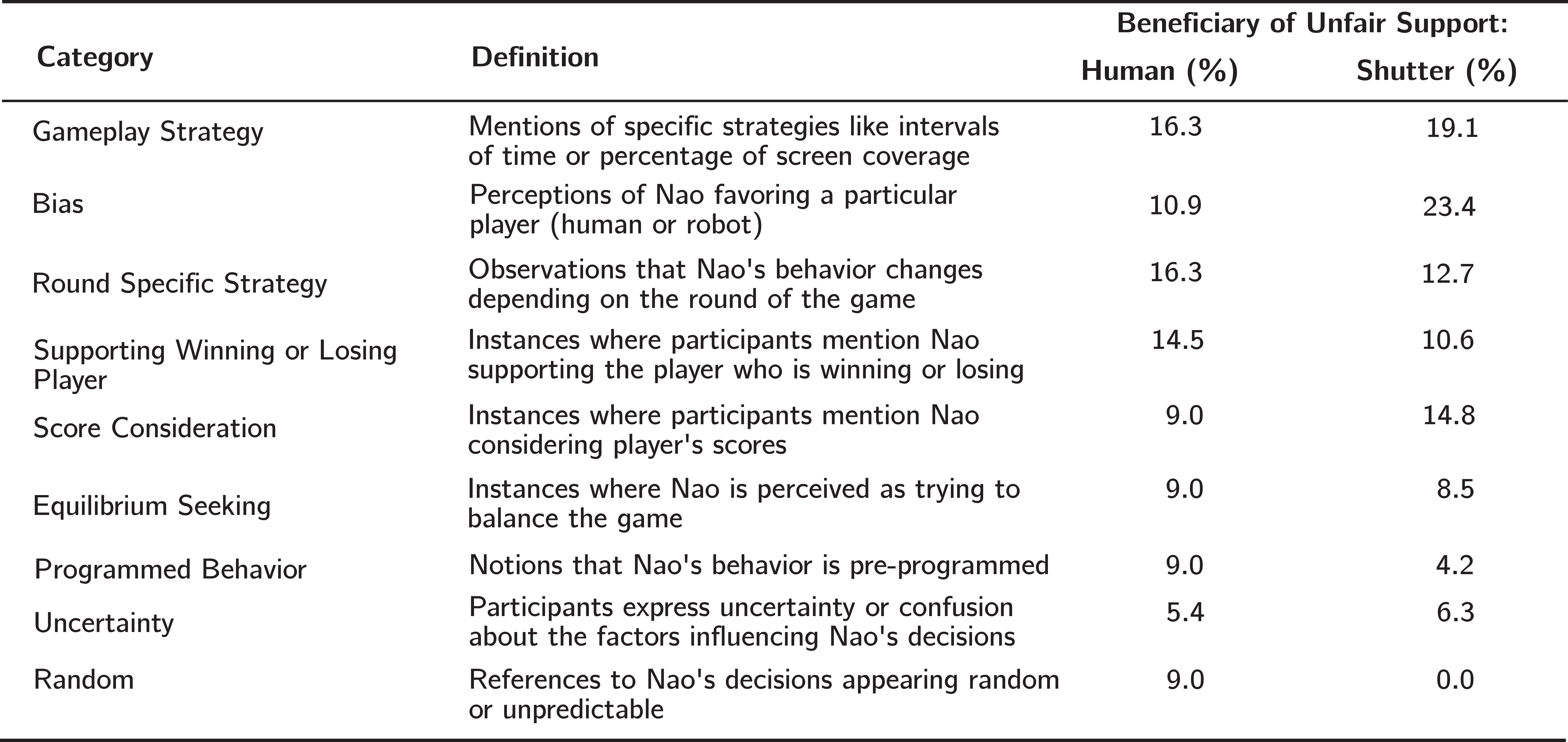}
\end{table}

\subsection{Perceived Closeness}

We found that the Beneficiary of Nao's Unfair Support had a significant effect on the perceived closeness between Nao and Shutter (\textit{F}[1,36] = 98.7, \textit{p} < .0001). As one would expect, participants felt that Nao and Shutter were significantly closer to one another when Shutter benefited from the unfair support compared to when the human did (\textit{p} $< .0001$). We also found a significant interaction effect of Game and Beneficiary of Unfair Support on the perceived closeness between Nao and Shutter (\textit{F}[2,74] = 66.5, \textit{p} < .0001). The post-hoc test indicated no significant differences on closeness 
for the games with equal support; however, for the other games, closeness perceptions for the two robots were significantly higher when Nao provided more support to Shutter.

In terms of the perceived closeness between the participants and the Nao robot, we found that the Beneficiary of Unfair Support had a significant effect on the ratings 
(\textit{F}[1,36] = 64.8, \textit{p}~$< .0001$). Participants felt significantly closer to Nao when they benefited from the unfair support compared to when Shutter benefited (p~$< 0.0001$). We also found a significant interaction effect of Game and Beneficiary of Unfair Support on these IOS ratings (\textit{F}[2,74] = 35.5, \textit{p} $< 0.0001$). These results were similar to the other IOS scale results, but higher closeness ratings were observed for the participants and Nao when the participants benefited from the help of this robot. 

\subsection{Trust}
 We found that the participants' moral trust toward Nao was significantly affected by the Game (\textit{F}[2,102]=8.32, p < .0001). The post-hoc tests showed that participants perceived higher moral trust toward Nao in the Equal Support Game (\textit{M}=4.21, \textit{SE}=.32) rather than in the Early Unfairness Game (\textit{M} = 2.55, \textit{SE}=.32) and in the Late Unfairness Game(\textit{M} = 2.64, \textit{SE} = .32). 

We found a significant effect of the game on performance trust ratings (\textit{F}[2,110] = 8.09, p = .0005). Participants reported higher levels of performance trust toward Nao when its support was equally distributed compared to other games the participants experienced (Early Unfairness: \textit{M} = 5.36, \textit{SE} = .20, \textit{p} = .0005; Late Unfairness: \textit{M} = 4.42, \textit{SE} = .20, \textit{p} = .004). The Beneficiary of Unfair Support (\textit{F}[2,110] = 8.09, \textit{p}=.0005) also had a significant impact on the performance trust ratings. Participants reported higher levels of the performance trust when they benefited from Nao’s unequal support (\textit{M} = 5.00, \textit{SE}= .16) than when Shutter benefited (\textit{M}= 4.38 , \textit{SE} = .16). 

As stated earlier, participants could select "NA" if they thought the suggested trust attribute was not applicable to Nao. Consistent with the approach of a previous study \cite{Tolmeijer2022}, we conducted a Student's t-test to compare the differences in the percentages of "NA" selections across the trust sub-scales.
We found a significant difference between the moral trust scales and the performance trust scales: participants selected “NA” options more frequently when rating moral trust toward Nao (\textit{M} = 41.4, \textit{SE} = 3.39, \textit{t} = 10.4, \textit{p} < .0001) compared to performance trust (\textit{M} = 5.20, \textit{SE} = .73). Further analysis showed that the average percentage of “NA” selections varied by the beneficiary of unfair support (\textit{t} = 2.71, \textit{p} = .0073) only when participants rated the moral trust. Those who received more support than Shutter were more likely to select "NA" options (\textit{M} = 50.4, \textit{SE} = 4.55) compared to those who received less support than Shutter (\textit{M} = 32.5, \textit{SE} = 4.77). 

\begin{figure}[t!p]
  \centering  \includegraphics[width=.40\linewidth]{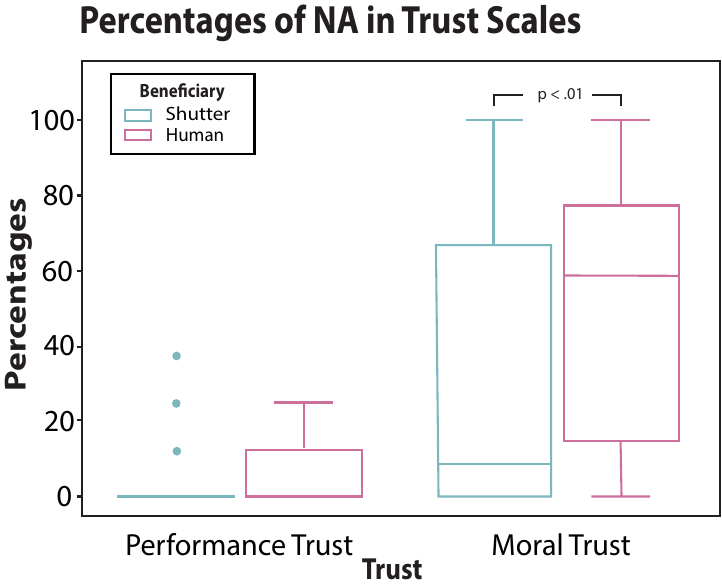}
    \caption{Average percentages of "Not Applicable"(NA) in Trust Scales. The boxes represent the interquartile ranges of ratings. The dots outside the boxes denote outliers.}
    \label{fig:trust_na}
\vspace{-1em}
\end{figure}
\subsection{Predicting Momentary Unfairness Perceptions based on the Fairness Components}
 Our fourth hypothesis (H4) stated that three components of fairness according to the Fairness Theory \cite{folger2001fairness} could be used to predict momentary unfairness perceptions throughout an interaction. 
 To evaluate this hypothesis, we exclusively considered responses from participants who experienced Shutter receiving unfair support. This focus was necessitated by the structure of the reduced welfare question, which is specifically designed to gauge the perception of reduced welfare from the participant's viewpoint. 
 
Initially, we conducted three correlation analyses considering  the responses to the questions about participant's perceptions of each factor of the Fairness Theory and perceptions of momentary unfairness (reversed ratings for momentary fairness, as in Sec. \ref{ssec:results_momentary_fairness}, on a 5 point scale). The responses to the moral transgression question (``How would you rate the social appropriateness
of the robot’s decisions in the context of the competition between yourself and Shutter?'') were reversed to align with the momentary unfairness ratings. We found that the responses to the reduced welfare ($r = .62, p < .0001 $), conduct ($r = .49, p < .0001 $), and moral transgression ($r = .80, p < .0001 $) questions all demonstrated moderate to strong positive correlation, suggesting that each factor could be relevant for predicting momentary unfairness perceptions. 
 
 We considered using a traditional statistical analysis with a linear model to predict momentary fairness ratings based on perceptions of the factors of the Fairness Theory; however, because we did not control explicitly for the three factors of the Fairness Theory, we opted instead for a simple classification approach through which we could analyze prediction performance without fitting any parameters and considering a variety of well-established performance metrics. Through this analysis, we evaluated how well momentary impressions of an unfair Nao robot (as binary values) could be predicted by the conjunction of boolean variables representing the welfare, conduct, and moral transgression factors:
\begin{align}
    \mathtt{unfair\_robot} = \mathtt{reduced\_welfare}\, \wedge \, \mathtt{conduct} \, \wedge \, \mathtt{moral\_transgression}
    \label{eq:unfair_classifier}
\end{align}



\setlength{\tabcolsep}{2.5pt}
\begin{table}[t]
\caption{Performance of boolean classifiers based on participants' perceptions of the components of the Fairness Theory \cite{folger2001fairness}. The results consider \textit{all} the data from the participants that experienced the Nao robot benefiting the Shutter robot more than them in the second and third game. We highlight best $F_1$ scores in bold, although these results may be obtained with a different tradeoff between precision and recall.}
{\footnotesize
\begin{tabular}{llccc}
& {\small Model} & {\small Recall} & {\small Precision} & {\small $F_1$} \\ \hline
\rownumber & $\texttt{reduced\_welfare}$  & 0.91 & 0.86 & \textbf{0.89} \\
\rownumber & $\texttt{conduct}$  & 0.72 & 0.81 &  0.76 \\
\rownumber & $\texttt{moral\_transgression}$ &  0.84 & 0.92 &  0.88 \\
\rownumber & $\texttt{reduced\_welfare} \wedge \texttt{conduct}$ &  0.70 & 1.00  &  0.82 \\
\rownumber & $\texttt{reduced\_welfare} \wedge \texttt{moral\_transgression}$ &  0.80 & 1.00 &  \textbf{0.89} \\
\rownumber & $\texttt{conduct} \wedge \texttt{moral\_transgression}$ &  0.67 & 1.00 &  0.80 \\
\rownumber & $\texttt{reduced\_welfare} \wedge \texttt{conduct} \wedge \texttt{moral\_transgression}$ &  0.65 & 0.98 & 0.78 \\

\end{tabular}
}
\label{tab:boolean}
\end{table}

\setlength{\tabcolsep}{3pt}
\begin{table}
\caption{Performance of boolean classifiers considering the data from the participants that experienced the Nao robot benefiting the Shutter robot more than them in the second and third game. The results are split by Game (Equal Support, Early Unfairness, Late Unfairness). Rec. stands for recall and Prec. stands for precision. $F_1$ is the harmonic mean of the precision and recall metrics. We highlight best $F_1$ scores in bold per game.}
{\footnotesize
\begin{tabular}{ll|ccc|ccc|ccc}
&  & \multicolumn{3}{c|}{\textit{Equal Support}} & \multicolumn{3}{c|}{\textit{Early Unfairness}} & \multicolumn{3}{c}{\textit{Late Unfairness}} \\
& Model & Rec. & Prec. & $F_1$ & Rec. & Prec. & $F_1$ & Rec. & Prec. & $F_1$ \\ \hline
\setcounter{rownum}{0}
\rownumber & $\texttt{reduced\_welfare}$ & 0.67 & 0.54 & 0.60 & 0.97 & 0.92 & 0.95 & 0.84 & 0.79 & \textbf{0.82} \\
\rownumber & $\texttt{conduct}$ & 0.53 & 0.44 &  0.48  & 0.78 & 0.91 & 0.84 & 0.66 & 0.70 & 0.68  \\
\rownumber & $\texttt{moral\_transgression}$ & 0.57 & 0.68 &  \textbf{0.62}  & 0.92 & 0.97 & 0.94 & 0.75 & 0.86 & 0.80 \\
\rownumber & $\texttt{reduced\_welfare} \wedge \texttt{conduct}$  & 0.70 & 0.24 &  0.36  & 0.78 & 1.00 & 0.88 & 0.59 & 1.00 & 0.75 \\
\rownumber & $\texttt{reduced\_welfare} \wedge \texttt{moral\_transgression}$  & 0.63 & 0.25 & 0.36  & 0.92 & 1.00 & \textbf{0.96} & 0.66 & 1.00 & 0.79 \\
\rownumber & $\texttt{conduct} \wedge \texttt{moral\_transgression}$ &  0.63 & 0.25 & 0.36  & 0.78 & 1.00 & 0.88 & 0.72 & 0.53 & 0.69 \\
\rownumber & $\texttt{reduced\_welfare} \wedge \texttt{conduct} \wedge \texttt{moral\_transgression}$  & 0.30 & 0.82 & 0.44  & 0.78 & 1.00 & 0.88 & 0.50 & 0.94 & 0.65 \\
\end{tabular}
}
\label{tab:boolean_Games}
\end{table}

\setlength{\tabcolsep}{3pt}
\begin{table}
\caption{Performance of boolean classifiers considering the data from the participants that experienced the Nao robot benefiting Shutter more than them in the second and third game. The results are split by the Order in which they  experienced the Early and Late Unfairness games. Rec. is recall and Prec. is precision. $F_1$ is the harmonic mean of the precision and recall metrics. We highlight best $F_1$ scores in bold for each Order.}
{\footnotesize
\begin{tabular}{ll|ccc|ccc}
&  & \multicolumn{3}{c|}{\textit{Early Unfairness First}} & \multicolumn{3}{c}{\textit{Late Unfairness First}} \\
& Model & Rec. & Prec. & $F_1$ & Rec. & Prec. & $F_1$  \\ \hline
\setcounter{rownum}{0}
\rownumber & $\texttt{reduced\_welfare}$ & 0.80 & 0.80 & 0.80 & 0.95 & 0.80 & \textbf{0.87}  \\
\rownumber & $\texttt{conduct}$ & 0.61 & 0.79 &  0.69  & 0.79 & 0.69 &  0.74  \\
\rownumber & $\texttt{moral\_transgression}$ & 0.82 & 0.97 &  \textbf{0.89}  & 0.74 & 0.83 & 0.78\\
\rownumber & $\texttt{reduced\_welfare} \wedge \texttt{conduct}$  & 0.57 & 0.96 &  0.71  & 0.77 & 0.91 & 0.83  \\
\rownumber & $\texttt{reduced\_welfare} \wedge \texttt{moral\_transgression}$  & 0.60 & 0.97 & 0.82  & 0.74 & 0.91 &  0.82\\
\rownumber & $\texttt{conduct} \wedge \texttt{moral\_transgression}$ &  0.59 & 1.00 & 0.74  & 0.64 & 0.83 & 0.72 \\
\rownumber & $\texttt{reduced\_welfare} \wedge \texttt{conduct} \wedge \texttt{moral\_transgression}$  & 0.57 & 1.00 & 0.72  & 0.64 & 0.93 & 0.76 \\
\vspace{-2em}

\end{tabular}
}
\label{tab:boolean_Order}
\end{table}

 \noindent
 The binary variables were computed as follows. First, 
 the ground truth values for  $\mathtt{unfair\_robot}$ in eq. (\ref{eq:unfair_classifier}) were computed using momentary fairness ratings: any value below 3 was recorded as 1 (indicating perceived unfairness), and all other values were recorded as 0. 
 Then, 
 for $\mathtt{reduced\_welfare}$  and $\mathtt{conduct}$ we used the ratings to ``Nao’s gameplay strategy significantly reduced my chances
of success'' and  ``How strongly do you
feel that the robot had feasible alternatives to its actions in this part
of the game?'', respectively. Any value above 3 was recorded as 1 (indicating the presence of the threat or potential alternative actions that Nao could have taken), and all other values were recorded as 0. 
For $\mathtt{moral\_transgression}$, any response below 3 to ``How would you rate the social appropriateness
of the robot’s decisions in the context of the competition between
yourself and Shutter?'' was recorded as 1 (indicating the presence of an ethical violation by Nao), while values of 3 or above were recorded as 0. 
Finally, we identified instances where the three fairness components ($\mathtt{reduced\_welfare}$, $\mathtt{conduct}$, and $\mathtt{moral\_transgression}$) were all true in eq. (\ref{eq:unfair_classifier}) 
and examined if they matched the  $\mathtt{unfair\_robot}$ ground truth values. 

Tables \ref{tab:boolean}, \ref{tab:boolean_Games} and \ref{tab:boolean_Order} show the performance of the model from eq. (\ref{eq:unfair_classifier}) in comparison to similar models that used fewer input variables. The tables report three classification metrics: \textit{precision} (the ratio of true positives over all samples predicted to be positive), \textit{recall} (the true positive rate, or the number of true positives divided by the number of all samples that should have been identified as positives), and the $F_1$ score (the harmonic mean between precision and recall, which symmetrically represents both precision and recall). Higher values in the tables are better. 

Results show that the best classification performance, considering the $F_1$ score, was obtained with a simpler classifier that included $\mathtt{reduced\_welfare}$, or $\mathtt{moral\_transgression}$, or a combination of both of these factors. This suggests that unfairness perceptions may be highly contextual, with some fairness components having a higher impact in some cases than others. Moreover, results seem to indicate that a simple conjunction of the factors is not an effective way to consider all three factors when predicting unfairness perceptions. At times, the conjunction of two strong factors lowers $F_1$ performance (for example, compare row 5 and row 1 for Late Unfairness games in Table \ref{tab:boolean_Games}). More research is needed to devise better mathematical models that can be used to predict unfairness perceptions based on the three components of the theory. We also see that $\mathtt{conduct}$ was often the worst predictor for unfairness perceptions. The  results with our chosen model from eq. (\ref{eq:unfair_classifier}) do not support our fourth hypothesis (H4).


To help us understand the above results, we further investigated whether  $\mathtt{conduct}$ did not play as important of a role as expected for predicting perceptions of unfair robot behavior because people may have perceived the robot as capable of taking different actions all throughout the games. This was indeed supported by our data as most participants agreed the robot had feasible action alternatives. Only 24\% of participants rated the conduct factor below 3 on a 5-point scale.

\section{Discussion}


Research into fairness in HRI has found that people are attuned to how they are treated relative to others by a robot \cite{claure2020multi,jung2020robot}. However, many of these findings focus on fairness perceptions at the end of an interaction. Our results extend this line of work by highlighting how participants' perceptions of fairness are updated as they observe a robot's actions and the consequences that come with the actions.

\subsection{Implications for Fairness in HRI}


For our first research question, we investigated how perceptions of fairness change over time given unfair actions from a robot, either early or late in the interaction. Our results partially support our first hypothesis, showing that perceptions of fairness decline early on when unfair support is introduced at the beginning and drop later when unfair support comes toward the end. These findings are in line with recent trends exploring fairness in AI algorithms \cite{claure2023social,d2020fairness} and organizational psychology \cite{jones2013perceptions}, which have suggested the idea that fairness perceptions can update over time. Moreover, our findings are consistent with research indicating that inconsistencies in treatment can lead to reduced perceptions of fairness \cite{matta2017consistently}. Interestingly, when unfairness by a robot was experienced early in a Space Invaders game in our study, perceptions of fairness towards the robot did not seem to recover over time. It would be interesting to explore this phenomenon in the future over longer human-robot interactions.

We also examined the influence of the timing of a robot's unfair actions on participants' overall perception of fairness. As expected, game rounds were rated most fair when the Nao robot equally distributed its support. Contrary to our expectation, however, we found no significant differences in overall fairness perceptions based on whether the unfair behavior occurred early or late in the interaction. The lack of such temporal effect could be due to the short duration of the Space Invaders games, or due to the video self-annotation process that took place before participants provided overall fairness ratings. The process reduced the need for memory recall and could have mitigated recency effects in the ratings \cite{steiner1989immediate}.


 For our third research question, we explored whether  fairness perceptions would be positively influenced if a human were the beneficiary of Nao's unfair actions. Momentary fairness ratings did not support this hypothesis. For overall fairness, we found a trend that suggested that Nao could be perceived as more fair considering the overall game when the participants received more help from it than their competitor. This was further supported by answers to our open-ended question, showing that participants viewed Nao's decisions as strategic when they gained from the unfair support, but saw them as biased when the support favored their competitor. Prior findings have found people have strong responses to unfair actions from AI systems \cite{claure2020multi} and it seems that human sensitivity to unfairness is robust.

Prior work shows that robots exhibiting unfair behavior can impact how we behave and work with one another \cite{jung2020robot,sakamoto2006sociality,erel2024rosi,gillet2024interaction}. 
This motivated us to explore the social consequences -- interpersonal perceptions and trust toward robots -- that interactions with an unfair robot might bring in a competitive context. We found that perceived closeness towards different members is contingent on who receives more support. Specifically, participants felt closer to Nao when it unfairly supported them. Conversely, they perceived a closer relationship between Nao and Shutter when Shutter received more support. This reinforces the growing body of work around the idea that unfair robot behavior can actively shape perceptions of group dynamics in HRI \cite{chang2021unfair,sebo2020robots}.

We further explored how the timing of unequal support and the recipient of unfair behaviors influenced performance and moral trust toward Nao. Our findings indicated that both moral and performance trust decreased if Nao provided unequal support. The timing of unequal support (early unequal support vs. late unequal support) had no significant impact on performance or moral trust. These results build on existing research highlighting the impact of disproportionate support by robots \cite{claure2020multi}, demonstrating that unfair actions toward humans diminish trust regardless of when the unfair actions occur.
The results on performance trust reflected the self-serving bias that has been shown in human-robot collaboration experiments~\cite{you2011robot,van2021think}. Participants who benefited from Nao's unequal support trusted a robot's performance more than those who did not benefit from the additional support. However, this effect did not extend to moral trust. 
Interestingly, we found that more than 30\% of participants believed the moral rating scales were “not applicable” to Nao. This finding may be related to people's tendency to take into account the intention and the context behind the actions when judging the morality of other people\cite{gray2007moral,falk2008testing}.  It is well established that people can perceive robots as social agents \cite{groom2007can}; however, the way in which the perceived agency of the robot influences how humans interpret moral decisions and intentions of the robot remains a topic of ongoing discussion \cite{frazier2022perceived}. Further research is needed to explore the interplay between trust, fairness, and the perceived agency of robots on moral judgments in human-robot interactions.

Our fourth research question pertained to the use of the Fairness Theory's \cite{folger2001fairness} key components -- reduced welfare, conduct, and moral transgression -- to predict momentary fairness in human-robot interactions. Our results suggest that fairness judgments were mainly based on perceived reduced welfare and moral violations by Nao rather than on all three factors. This raises interesting questions about how the nuances of interacting with a robot may affect the fairness components. It may be that the conduct factor does not translate easily from human-human interactions to HRI because it requires people to perceive the robot as having the capacity to make autonomous choices that are genuinely their own. To explore this further, future work should investigate the degree to which participants view the robot as having the agency to make its own decisions. Moreover, the competitive context in our study could have shaped our results. It would be interesting to investigate in future work the importance of the fairness factors in alternative HRI scenarios, e.g., search and rescue or other resource allocation tasks  \cite{brandao2020fair,claure2020multi}. Last but not least, it would be interesting to extend our analysis of the predictive power of the Fairness Theory from a feature ablation study to a more explicit causal analysis. This would require controlling for reduced welfare, conduct, and moral transgression perceptions towards robots in future studies.

 \subsection{Limitations and Future Work}
Our work is not without limitations. First, we studied fairness perceptions in a competitive multi-player Space Invaders game. In the future, it would be interesting to study fairness  on a moment-to-moment basis in other competitive settings or in fully collaborative interaction contexts, including problem solving tasks \cite{tennent2019micbot} and learning contexts \cite{gillet2022learning}. Second, the participant's opponent in the Space Invaders game was a robot. This brings up the question of what would happen if the competitor was a different robot (e.g., more or less anthropomorphic) or even another human. Finally, the human-robot interactions studied in this work were short and we measured fairness only at three points in time during a Space Invaders game. Future work should investigate moment-to-moment fairness perceptions and the application of the Fairness Theory \cite{folger2001fairness} to longer-term HRI scenarios. Additionally, future work could explore other methodologies to capture fairness perceptions in addition to video surveys, e.g., perhaps a robot can ask a human directly what they think about its  behavior, or augmented reality interfaces \cite{suzuki2022augmented} could be used to query humans about their perceptions of a robot as interactions evolve  rather  than querying them after they end. 

\section{Conclusion}
Robots are increasingly being placed in situations where they have to make decisions that raise fairness concerns from humans (e.g. deciding which tasks to allocate to team members \cite{gombolay2015decision} or making decisions on who to prioritize during a search and rescue context \cite{brandao2020fair}). In order to build appropriate algorithms that will ensure that these robots follow human notions of fairness, it is vital to understand how fairness judgments by humans evolve during interactions with robots. In this work, we found that fairness is not static. When people analyze an unfair action from a robot, the timing of an unfair action is a factor that updates their fairness perceptions. We further explored using the Fairness Theory \cite{folger2001fairness} to predict momentary fairness perceptions, and found that two of its components (reduced welfare and moral transgression) seemed to be better predictors than all three components. This could have been due to the competitive nature of human-robot interactions in our study. Finally, our work showed that a robot's unfair support can shape trust toward a robot and group perceptions within a competitive context. Our results are an initial step in understanding the dynamics of fairness in HRI.


\bibliographystyle{ACM-Reference-Format}
\bibliography{references.bib}


\begin{thebibliography}{96}


\ifx \showCODEN    \undefined \def \showCODEN     #1{\unskip}     \fi
\ifx \showDOI      \undefined \def \showDOI       #1{#1}\fi
\ifx \showISBNx    \undefined \def \showISBNx     #1{\unskip}     \fi
\ifx \showISBNxiii \undefined \def \showISBNxiii  #1{\unskip}     \fi
\ifx \showISSN     \undefined \def \showISSN      #1{\unskip}     \fi
\ifx \showLCCN     \undefined \def \showLCCN      #1{\unskip}     \fi
\ifx \shownote     \undefined \def \shownote      #1{#1}          \fi
\ifx \showarticletitle \undefined \def \showarticletitle #1{#1}   \fi
\ifx \showURL      \undefined \def \showURL       {\relax}        \fi
\providecommand\bibfield[2]{#2}
\providecommand\bibinfo[2]{#2}
\providecommand\natexlab[1]{#1}
\providecommand\showeprint[2][]{arXiv:#2}

\bibitem[Adams(1965)]%
        {adams1965inequity}
\bibfield{author}{\bibinfo{person}{J~Stacy Adams}.} \bibinfo{year}{1965}\natexlab{}.
\newblock \showarticletitle{Inequity in social exchange}.
\newblock In \bibinfo{booktitle}{\emph{Advances in experimental social psychology}}. Vol.~\bibinfo{volume}{2}. \bibinfo{publisher}{Elsevier}, \bibinfo{pages}{267--299}.
\newblock


\bibitem[Adamson et~al\mbox{.}(2020)]%
        {adamson2020designing}
\bibfield{author}{\bibinfo{person}{Timothy Adamson}, \bibinfo{person}{C~Burton Lyng-Olsen}, \bibinfo{person}{Kendrick Umstattd}, {and} \bibinfo{person}{Marynel V{\'a}zquez}.} \bibinfo{year}{2020}\natexlab{}.
\newblock \showarticletitle{Designing social interactions with a humorous robot photographer}. In \bibinfo{booktitle}{\emph{Proceedings of the 2020 ACM/IEEE International Conference on Human-Robot Interaction}}. \bibinfo{pages}{233--241}.
\newblock


\bibitem[Arnold and Scheutz(2018)]%
        {arnold2018observing}
\bibfield{author}{\bibinfo{person}{Thomas Arnold} {and} \bibinfo{person}{Matthias Scheutz}.} \bibinfo{year}{2018}\natexlab{}.
\newblock \showarticletitle{Observing robot touch in context: How does touch and attitude affect perceptions of a robot’s social qualities?}. In \bibinfo{booktitle}{\emph{2018 13th ACM/IEEE International Conference on Human-Robot Interaction (HRI)}}. IEEE, \bibinfo{pages}{352--360}.
\newblock


\bibitem[Aron et~al\mbox{.}(1992)]%
        {aron1992inclusion}
\bibfield{author}{\bibinfo{person}{Arthur Aron}, \bibinfo{person}{Elaine~N Aron}, {and} \bibinfo{person}{Danny Smollan}.} \bibinfo{year}{1992}\natexlab{}.
\newblock \showarticletitle{Inclusion of other in the self scale and the structure of interpersonal closeness.}
\newblock \bibinfo{journal}{\emph{Journal of personality and social psychology}} \bibinfo{volume}{63}, \bibinfo{number}{4} (\bibinfo{year}{1992}), \bibinfo{pages}{596}.
\newblock


\bibitem[Barocas et~al\mbox{.}(2017)]%
        {barocas2017fairness}
\bibfield{author}{\bibinfo{person}{Solon Barocas}, \bibinfo{person}{Moritz Hardt}, {and} \bibinfo{person}{Arvind Narayanan}.} \bibinfo{year}{2017}\natexlab{}.
\newblock \showarticletitle{Fairness in machine learning}.
\newblock \bibinfo{journal}{\emph{Nips tutorial}}  \bibinfo{volume}{1} (\bibinfo{year}{2017}), \bibinfo{pages}{2}.
\newblock


\bibitem[Barocas and Selbst(2016)]%
        {barocas2016big}
\bibfield{author}{\bibinfo{person}{Solon Barocas} {and} \bibinfo{person}{Andrew~D Selbst}.} \bibinfo{year}{2016}\natexlab{}.
\newblock \showarticletitle{Big data's disparate impact}.
\newblock \bibinfo{journal}{\emph{Calif. L. Rev.}}  \bibinfo{volume}{104} (\bibinfo{year}{2016}), \bibinfo{pages}{671}.
\newblock


\bibitem[Bartneck et~al\mbox{.}(2018)]%
        {bartneck2018robots}
\bibfield{author}{\bibinfo{person}{Christoph Bartneck}, \bibinfo{person}{Kumar Yogeeswaran}, \bibinfo{person}{Qi~Min Ser}, \bibinfo{person}{Graeme Woodward}, \bibinfo{person}{Robert Sparrow}, \bibinfo{person}{Siheng Wang}, {and} \bibinfo{person}{Friederike Eyssel}.} \bibinfo{year}{2018}\natexlab{}.
\newblock \showarticletitle{Robots and racism}. In \bibinfo{booktitle}{\emph{Proceedings of the 2018 ACM/IEEE international conference on human-robot interaction}}. \bibinfo{pages}{196--204}.
\newblock


\bibitem[Bartol and Srivastava(2002)]%
        {bartol2002encouraging}
\bibfield{author}{\bibinfo{person}{Kathryn~M Bartol} {and} \bibinfo{person}{Abhishek Srivastava}.} \bibinfo{year}{2002}\natexlab{}.
\newblock \showarticletitle{Encouraging knowledge sharing: The role of organizational reward systems}.
\newblock \bibinfo{journal}{\emph{Journal of leadership \& organizational studies}} \bibinfo{volume}{9}, \bibinfo{number}{1} (\bibinfo{year}{2002}), \bibinfo{pages}{64--76}.
\newblock


\bibitem[Bettencourt and Brown(1997)]%
        {bettencourt1997contact}
\bibfield{author}{\bibinfo{person}{Lance~A Bettencourt} {and} \bibinfo{person}{Stephen~W Brown}.} \bibinfo{year}{1997}\natexlab{}.
\newblock \showarticletitle{Contact employees: Relationships among workplace fairness, job satisfaction and prosocial service behaviors}.
\newblock \bibinfo{journal}{\emph{Journal of retailing}} \bibinfo{volume}{73}, \bibinfo{number}{1} (\bibinfo{year}{1997}), \bibinfo{pages}{39--61}.
\newblock


\bibitem[Brand{\~a}o(2021)]%
        {brandao2021socially}
\bibfield{author}{\bibinfo{person}{Martim Brand{\~a}o}.} \bibinfo{year}{2021}\natexlab{}.
\newblock \showarticletitle{Socially fair coverage: The fairness problem in coverage planning and a new anytime-fair method}. In \bibinfo{booktitle}{\emph{2021 IEEE International Conference on Advanced Robotics and Its Social Impacts (ARSO)}}. IEEE, \bibinfo{pages}{227--233}.
\newblock


\bibitem[Brandao et~al\mbox{.}(2020)]%
        {brandao2020fair}
\bibfield{author}{\bibinfo{person}{Martim Brandao}, \bibinfo{person}{Marina Jirotka}, \bibinfo{person}{Helena Webb}, {and} \bibinfo{person}{Paul Luff}.} \bibinfo{year}{2020}\natexlab{}.
\newblock \showarticletitle{Fair navigation planning: a resource for characterizing and designing fairness in mobile robots}.
\newblock \bibinfo{journal}{\emph{Artificial Intelligence}}  \bibinfo{volume}{282} (\bibinfo{year}{2020}), \bibinfo{pages}{103259}.
\newblock


\bibitem[Buolamwini and Gebru(2018)]%
        {buolamwini2018gender}
\bibfield{author}{\bibinfo{person}{Joy Buolamwini} {and} \bibinfo{person}{Timnit Gebru}.} \bibinfo{year}{2018}\natexlab{}.
\newblock \showarticletitle{Gender shades: Intersectional accuracy disparities in commercial gender classification}. In \bibinfo{booktitle}{\emph{Conference on fairness, accountability and transparency}}. PMLR, \bibinfo{pages}{77--91}.
\newblock


\bibitem[Candon et~al\mbox{.}(2022)]%
        {candon2022perceptions}
\bibfield{author}{\bibinfo{person}{Kate Candon}, \bibinfo{person}{Zoe Hsu}, \bibinfo{person}{Yoony Kim}, \bibinfo{person}{Jesse Chen}, \bibinfo{person}{Nathan Tsoi}, {and} \bibinfo{person}{Marynel V{\'a}zquez}.} \bibinfo{year}{2022}\natexlab{}.
\newblock \showarticletitle{Perceptions of the Helpfulness of Unexpected Agent Assistance}. In \bibinfo{booktitle}{\emph{Proceedings of the 10th International Conference on Human-Agent Interaction}}. \bibinfo{pages}{41--50}.
\newblock


\bibitem[Candon et~al\mbox{.}(2023)]%
        {candon2023verbally}
\bibfield{author}{\bibinfo{person}{Kate Candon}, \bibinfo{person}{Helen Zhou}, \bibinfo{person}{Sarah Gillet}, {and} \bibinfo{person}{Marynel V{\'a}zquez}.} \bibinfo{year}{2023}\natexlab{}.
\newblock \showarticletitle{Verbally Soliciting Human Feedback in Continuous Human-Robot Collaboration: Effects of the Framing and Timing of Reminders}. In \bibinfo{booktitle}{\emph{Proceedings of the 2023 ACM/IEEE International Conference on Human-Robot Interaction}}. \bibinfo{pages}{290--300}.
\newblock


\bibitem[Chang et~al\mbox{.}(2020)]%
        {chang2020defining}
\bibfield{author}{\bibinfo{person}{Mai~Lee Chang}, \bibinfo{person}{Zachary Pope}, \bibinfo{person}{Elaine~Schaertl Short}, {and} \bibinfo{person}{Andrea~Lockerd Thomaz}.} \bibinfo{year}{2020}\natexlab{}.
\newblock \showarticletitle{Defining fairness in human-robot teams}. In \bibinfo{booktitle}{\emph{2020 29th IEEE International Conference on Robot and Human Interactive Communication (RO-MAN)}}. IEEE, \bibinfo{pages}{1251--1258}.
\newblock


\bibitem[Chang et~al\mbox{.}(2021)]%
        {chang2021unfair}
\bibfield{author}{\bibinfo{person}{Mai~Lee Chang}, \bibinfo{person}{Greg Trafton}, \bibinfo{person}{J~Malcolm McCurry}, {and} \bibinfo{person}{Andrea~Lockerd Thomaz}.} \bibinfo{year}{2021}\natexlab{}.
\newblock \showarticletitle{Unfair! Perceptions of Fairness in Human-Robot Teams}. In \bibinfo{booktitle}{\emph{2021 30th IEEE International Conference on Robot \& Human Interactive Communication (RO-MAN)}}. IEEE, \bibinfo{pages}{905--912}.
\newblock


\bibitem[Chen et~al\mbox{.}(2018)]%
        {chen2018planning}
\bibfield{author}{\bibinfo{person}{Min Chen}, \bibinfo{person}{Stefanos Nikolaidis}, \bibinfo{person}{Harold Soh}, \bibinfo{person}{David Hsu}, {and} \bibinfo{person}{Siddhartha Srinivasa}.} \bibinfo{year}{2018}\natexlab{}.
\newblock \showarticletitle{Planning with trust for human-robot collaboration}. In \bibinfo{booktitle}{\emph{Proceedings of the 2018 ACM/IEEE international conference on human-robot interaction}}. \bibinfo{pages}{307--315}.
\newblock


\bibitem[Chi and Malle(2023)]%
        {chi2023people}
\bibfield{author}{\bibinfo{person}{Vivienne~Bihe Chi} {and} \bibinfo{person}{Bertram~F Malle}.} \bibinfo{year}{2023}\natexlab{}.
\newblock \showarticletitle{People dynamically update trust when interactively teaching robots}. In \bibinfo{booktitle}{\emph{Proceedings of the 2023 ACM/IEEE International Conference on Human-Robot Interaction. HRI}}, Vol.~\bibinfo{volume}{23}. \bibinfo{pages}{554--564}.
\newblock


\bibitem[Civai et~al\mbox{.}(2010)]%
        {civai2010irrational}
\bibfield{author}{\bibinfo{person}{Claudia Civai}, \bibinfo{person}{Corrado Corradi-Dell’Acqua}, \bibinfo{person}{Matthias Gamer}, {and} \bibinfo{person}{Raffaella~I Rumiati}.} \bibinfo{year}{2010}\natexlab{}.
\newblock \showarticletitle{Are irrational reactions to unfairness truly emotionally-driven? Dissociated behavioural and emotional responses in the Ultimatum Game task}.
\newblock \bibinfo{journal}{\emph{Cognition}} \bibinfo{volume}{114}, \bibinfo{number}{1} (\bibinfo{year}{2010}), \bibinfo{pages}{89--95}.
\newblock


\bibitem[Claure(2024)]%
        {claure2024designing}
\bibfield{author}{\bibinfo{person}{Houston Claure}.} \bibinfo{year}{2024}\natexlab{}.
\newblock \showarticletitle{Designing for Fairness in Human-Robot Interactions}.
\newblock \bibinfo{journal}{\emph{arXiv preprint arXiv:2405.21044}} (\bibinfo{year}{2024}).
\newblock


\bibitem[Claure et~al\mbox{.}({[n.\,d.]})]%
        {ClaureReinforcementTeams}
\bibfield{author}{\bibinfo{person}{Houston Claure}, \bibinfo{person}{Yifang Chen}, \bibinfo{person}{Jignesh Modi}, \bibinfo{person}{Malte Jung}, {and} \bibinfo{person}{Stefanos Nikolaidis}.} \bibinfo{year}{[n.\,d.]}\natexlab{}.
\newblock \bibinfo{booktitle}{\emph{{Reinforcement Learning with Fairness Constraints for Resource Distribution in Human-Robot Teams}}}.
\newblock \bibinfo{type}{{T}echnical {R}eport}.
\newblock


\bibitem[Claure et~al\mbox{.}(2020)]%
        {claure2020multi}
\bibfield{author}{\bibinfo{person}{Houston Claure}, \bibinfo{person}{Yifang Chen}, \bibinfo{person}{Jignesh Modi}, \bibinfo{person}{Malte Jung}, {and} \bibinfo{person}{Stefanos Nikolaidis}.} \bibinfo{year}{2020}\natexlab{}.
\newblock \showarticletitle{Multi-armed bandits with fairness constraints for distributing resources to human teammates}. In \bibinfo{booktitle}{\emph{Proceedings of the 2020 ACM/IEEE International Conference on Human-Robot Interaction}}. \bibinfo{pages}{299--308}.
\newblock


\bibitem[Claure et~al\mbox{.}(2023)]%
        {claure2023social}
\bibfield{author}{\bibinfo{person}{Houston Claure}, \bibinfo{person}{Seyun Kim}, \bibinfo{person}{Ren{\'e}~F Kizilcec}, {and} \bibinfo{person}{Malte Jung}.} \bibinfo{year}{2023}\natexlab{}.
\newblock \showarticletitle{The social consequences of machine allocation behavior: Fairness, interpersonal perceptions and performance}.
\newblock \bibinfo{journal}{\emph{Computers in Human Behavior}}  \bibinfo{volume}{146} (\bibinfo{year}{2023}), \bibinfo{pages}{107628}.
\newblock


\bibitem[Colledanchise and {\"O}gren(2018)]%
        {colledanchise2018behavior}
\bibfield{author}{\bibinfo{person}{Michele Colledanchise} {and} \bibinfo{person}{Petter {\"O}gren}.} \bibinfo{year}{2018}\natexlab{}.
\newblock \bibinfo{booktitle}{\emph{Behavior trees in robotics and AI: An introduction}}.
\newblock \bibinfo{publisher}{CRC Press}.
\newblock


\bibitem[Colquitt and Zipay(2015)]%
        {colquitt2015justice}
\bibfield{author}{\bibinfo{person}{Jason~A Colquitt} {and} \bibinfo{person}{Kate~P Zipay}.} \bibinfo{year}{2015}\natexlab{}.
\newblock \showarticletitle{Justice, fairness, and employee reactions}.
\newblock \bibinfo{journal}{\emph{Annu. Rev. Organ. Psychol. Organ. Behav.}} \bibinfo{volume}{2}, \bibinfo{number}{1} (\bibinfo{year}{2015}), \bibinfo{pages}{75--99}.
\newblock


\bibitem[Corbeil and Searle(1976)]%
        {corbeil1976restricted}
\bibfield{author}{\bibinfo{person}{Robert~R Corbeil} {and} \bibinfo{person}{Shayle~R Searle}.} \bibinfo{year}{1976}\natexlab{}.
\newblock \showarticletitle{Restricted maximum likelihood (REML) estimation of variance components in the mixed model}.
\newblock \bibinfo{journal}{\emph{Technometrics}} \bibinfo{volume}{18}, \bibinfo{number}{1} (\bibinfo{year}{1976}), \bibinfo{pages}{31--38}.
\newblock


\bibitem[D'Amour et~al\mbox{.}(2020)]%
        {d2020fairness}
\bibfield{author}{\bibinfo{person}{Alexander D'Amour}, \bibinfo{person}{Hansa Srinivasan}, \bibinfo{person}{James Atwood}, \bibinfo{person}{Pallavi Baljekar}, \bibinfo{person}{D Sculley}, {and} \bibinfo{person}{Yoni Halpern}.} \bibinfo{year}{2020}\natexlab{}.
\newblock \showarticletitle{Fairness is not static: deeper understanding of long term fairness via simulation studies}. In \bibinfo{booktitle}{\emph{Proceedings of the 2020 Conference on Fairness, Accountability, and Transparency}}. \bibinfo{pages}{525--534}.
\newblock


\bibitem[Dastin(2018)]%
        {dastin2018amazon}
\bibfield{author}{\bibinfo{person}{Jeffrey Dastin}.} \bibinfo{year}{2018}\natexlab{}.
\newblock \showarticletitle{Amazon scraps secret AI recruiting tool that showed bias against women}.
\newblock In \bibinfo{booktitle}{\emph{Ethics of Data and Analytics}}. \bibinfo{publisher}{Auerbach Publications}, \bibinfo{pages}{296--299}.
\newblock


\bibitem[Datta et~al\mbox{.}(2014)]%
        {datta2014automated}
\bibfield{author}{\bibinfo{person}{Amit Datta}, \bibinfo{person}{Michael~Carl Tschantz}, {and} \bibinfo{person}{Anupam Datta}.} \bibinfo{year}{2014}\natexlab{}.
\newblock \showarticletitle{Automated experiments on ad privacy settings: A tale of opacity, choice, and discrimination}.
\newblock \bibinfo{journal}{\emph{arXiv preprint arXiv:1408.6491}} (\bibinfo{year}{2014}).
\newblock


\bibitem[De~Cremer and Tyler(2007)]%
        {de2007effects}
\bibfield{author}{\bibinfo{person}{David De~Cremer} {and} \bibinfo{person}{Tom~R Tyler}.} \bibinfo{year}{2007}\natexlab{}.
\newblock \showarticletitle{The effects of trust in authority and procedural fairness on cooperation.}
\newblock \bibinfo{journal}{\emph{Journal of applied psychology}} \bibinfo{volume}{92}, \bibinfo{number}{3} (\bibinfo{year}{2007}), \bibinfo{pages}{639}.
\newblock


\bibitem[Dulebohn and Martocchio(1998)]%
        {dulebohn1998employee}
\bibfield{author}{\bibinfo{person}{James~H Dulebohn} {and} \bibinfo{person}{Joseph~J Martocchio}.} \bibinfo{year}{1998}\natexlab{}.
\newblock \showarticletitle{Employee perceptions of the fairness of work group incentive pay plans}.
\newblock \bibinfo{journal}{\emph{Journal of Management}} \bibinfo{volume}{24}, \bibinfo{number}{4} (\bibinfo{year}{1998}), \bibinfo{pages}{469--488}.
\newblock


\bibitem[Erel et~al\mbox{.}(2024)]%
        {erel2024rosi}
\bibfield{author}{\bibinfo{person}{Hadas Erel}, \bibinfo{person}{Marynel V{\'a}zquez}, \bibinfo{person}{Sarah Sebo}, \bibinfo{person}{Nicole Salomons}, \bibinfo{person}{Sarah Gillet}, {and} \bibinfo{person}{Brian Scassellati}.} \bibinfo{year}{2024}\natexlab{}.
\newblock \showarticletitle{RoSI: A Model for Predicting Robot Social Influence}.
\newblock \bibinfo{journal}{\emph{ACM Transactions on Human-Robot Interaction}} (\bibinfo{year}{2024}).
\newblock


\bibitem[Falk et~al\mbox{.}(2008)]%
        {falk2008testing}
\bibfield{author}{\bibinfo{person}{Armin Falk}, \bibinfo{person}{Ernst Fehr}, {and} \bibinfo{person}{Urs Fischbacher}.} \bibinfo{year}{2008}\natexlab{}.
\newblock \showarticletitle{Testing theories of fairness—Intentions matter}.
\newblock \bibinfo{journal}{\emph{Games and Economic Behavior}} \bibinfo{volume}{62}, \bibinfo{number}{1} (\bibinfo{year}{2008}), \bibinfo{pages}{287--303}.
\newblock


\bibitem[Fehr and Schmidt(1999)]%
        {fehr1999theory}
\bibfield{author}{\bibinfo{person}{Ernst Fehr} {and} \bibinfo{person}{Klaus~M Schmidt}.} \bibinfo{year}{1999}\natexlab{}.
\newblock \showarticletitle{A theory of fairness, competition, and cooperation}.
\newblock \bibinfo{journal}{\emph{The quarterly journal of economics}} \bibinfo{volume}{114}, \bibinfo{number}{3} (\bibinfo{year}{1999}), \bibinfo{pages}{817--868}.
\newblock


\bibitem[Folger and Cropanzano(2001)]%
        {folger2001fairness}
\bibfield{author}{\bibinfo{person}{Robert Folger} {and} \bibinfo{person}{Russell Cropanzano}.} \bibinfo{year}{2001}\natexlab{}.
\newblock \showarticletitle{Fairness theory: Justice as accountability}.
\newblock \bibinfo{journal}{\emph{Advances in organizational justice}} \bibinfo{volume}{1}, \bibinfo{number}{1-55} (\bibinfo{year}{2001}), \bibinfo{pages}{12}.
\newblock


\bibitem[Fraune(2020)]%
        {fraune2020our}
\bibfield{author}{\bibinfo{person}{Marlena~R Fraune}.} \bibinfo{year}{2020}\natexlab{}.
\newblock \showarticletitle{Our robots, our team: Robot anthropomorphism moderates group effects in human--robot teams}.
\newblock \bibinfo{journal}{\emph{Frontiers in psychology}}  \bibinfo{volume}{11} (\bibinfo{year}{2020}), \bibinfo{pages}{1275}.
\newblock


\bibitem[Fraune et~al\mbox{.}(2019)]%
        {fraune2019human}
\bibfield{author}{\bibinfo{person}{Marlena~R Fraune}, \bibinfo{person}{Steven Sherrin}, \bibinfo{person}{Selma {\v{S}}abanovi{\'c}}, {and} \bibinfo{person}{Eliot~R Smith}.} \bibinfo{year}{2019}\natexlab{}.
\newblock \showarticletitle{Is human-robot interaction more competitive between groups than between individuals?}. In \bibinfo{booktitle}{\emph{2019 14th ACM/IEEE International Conference on Human-Robot Interaction (HRI)}}. IEEE, \bibinfo{pages}{104--113}.
\newblock


\bibitem[Frazier-Young et~al\mbox{.}(2022)]%
        {frazier2022perceived}
\bibfield{author}{\bibinfo{person}{Chelsea Frazier-Young}, \bibinfo{person}{Malcolm McCurry}, \bibinfo{person}{Kevin Zish}, {and} \bibinfo{person}{Greg Trafton}.} \bibinfo{year}{2022}\natexlab{}.
\newblock \showarticletitle{Perceived agency changes performance and moral trust in robots}. In \bibinfo{booktitle}{\emph{Proceedings of the Annual Meeting of the Cognitive Science Society}}, Vol.~\bibinfo{volume}{44}.
\newblock


\bibitem[Gillet et~al\mbox{.}(2022)]%
        {gillet2022learning}
\bibfield{author}{\bibinfo{person}{Sarah Gillet}, \bibinfo{person}{Maria~Teresa Parreira}, \bibinfo{person}{Marynel V{\'a}zquez}, {and} \bibinfo{person}{Iolanda Leite}.} \bibinfo{year}{2022}\natexlab{}.
\newblock \showarticletitle{Learning gaze behaviors for balancing participation in group human-robot interactions}. In \bibinfo{booktitle}{\emph{2022 17th ACM/IEEE International Conference on Human-Robot Interaction (HRI)}}. IEEE, \bibinfo{pages}{265--274}.
\newblock


\bibitem[Gillet et~al\mbox{.}(2024)]%
        {gillet2024interaction}
\bibfield{author}{\bibinfo{person}{Sarah Gillet}, \bibinfo{person}{Marynel V{\'a}zquez}, \bibinfo{person}{Sean Andrist}, \bibinfo{person}{Iolanda Leite}, {and} \bibinfo{person}{Sarah Sebo}.} \bibinfo{year}{2024}\natexlab{}.
\newblock \showarticletitle{Interaction-Shaping Robotics: Robots that Influence Interactions between Other Agents}.
\newblock \bibinfo{journal}{\emph{ACM Transactions on Human-Robot Interaction}} (\bibinfo{year}{2024}).
\newblock


\bibitem[Gombolay et~al\mbox{.}(2015)]%
        {gombolay2015decision}
\bibfield{author}{\bibinfo{person}{Matthew~C Gombolay}, \bibinfo{person}{Reymundo~A Gutierrez}, \bibinfo{person}{Shanelle~G Clarke}, \bibinfo{person}{Giancarlo~F Sturla}, {and} \bibinfo{person}{Julie~A Shah}.} \bibinfo{year}{2015}\natexlab{}.
\newblock \showarticletitle{Decision-making authority, team efficiency and human worker satisfaction in mixed human--robot teams}.
\newblock \bibinfo{journal}{\emph{Autonomous Robots}} \bibinfo{volume}{39}, \bibinfo{number}{3} (\bibinfo{year}{2015}), \bibinfo{pages}{293--312}.
\newblock


\bibitem[Gray et~al\mbox{.}(2007)]%
        {gray2007moral}
\bibfield{author}{\bibinfo{person}{Heather~M Gray}, \bibinfo{person}{Kurt Gray}, {and} \bibinfo{person}{Daniel~M Wegner}.} \bibinfo{year}{2007}\natexlab{}.
\newblock \showarticletitle{Dimensions of mind perception}.
\newblock \bibinfo{journal}{\emph{science}} \bibinfo{volume}{315}, \bibinfo{number}{5812} (\bibinfo{year}{2007}), \bibinfo{pages}{619--619}.
\newblock


\bibitem[Groom and Nass(2007)]%
        {groom2007can}
\bibfield{author}{\bibinfo{person}{Victoria Groom} {and} \bibinfo{person}{Clifford Nass}.} \bibinfo{year}{2007}\natexlab{}.
\newblock \showarticletitle{Can robots be teammates?: Benchmarks in human--robot teams}.
\newblock \bibinfo{journal}{\emph{Interaction studies}} \bibinfo{volume}{8}, \bibinfo{number}{3} (\bibinfo{year}{2007}), \bibinfo{pages}{483--500}.
\newblock


\bibitem[Guo and Yang(2021)]%
        {guo2021modeling}
\bibfield{author}{\bibinfo{person}{Yaohui Guo} {and} \bibinfo{person}{X~Jessie Yang}.} \bibinfo{year}{2021}\natexlab{}.
\newblock \showarticletitle{Modeling and predicting trust dynamics in human--robot teaming: A Bayesian inference approach}.
\newblock \bibinfo{journal}{\emph{International Journal of Social Robotics}} \bibinfo{volume}{13}, \bibinfo{number}{8} (\bibinfo{year}{2021}), \bibinfo{pages}{1899--1909}.
\newblock


\bibitem[Halevy et~al\mbox{.}(2021)]%
        {halevy2021mitigating}
\bibfield{author}{\bibinfo{person}{Matan Halevy}, \bibinfo{person}{Camille Harris}, \bibinfo{person}{Amy Bruckman}, \bibinfo{person}{Diyi Yang}, {and} \bibinfo{person}{Ayanna Howard}.} \bibinfo{year}{2021}\natexlab{}.
\newblock \showarticletitle{Mitigating racial biases in toxic language detection with an equity-based ensemble framework}.
\newblock In \bibinfo{booktitle}{\emph{Equity and Access in Algorithms, Mechanisms, and Optimization}}. \bibinfo{pages}{1--11}.
\newblock


\bibitem[Hamann et~al\mbox{.}(2014)]%
        {hamann2014meritocratic}
\bibfield{author}{\bibinfo{person}{Katharina Hamann}, \bibinfo{person}{Johanna Bender}, {and} \bibinfo{person}{Michael Tomasello}.} \bibinfo{year}{2014}\natexlab{}.
\newblock \showarticletitle{Meritocratic sharing is based on collaboration in 3-year-olds.}
\newblock \bibinfo{journal}{\emph{Developmental Psychology}} \bibinfo{volume}{50}, \bibinfo{number}{1} (\bibinfo{year}{2014}), \bibinfo{pages}{121}.
\newblock


\bibitem[Ho et~al\mbox{.}(2020)]%
        {ho2020achieving}
\bibfield{author}{\bibinfo{person}{Seng-Beng Ho}, \bibinfo{person}{Xiwen Yang}, {and} \bibinfo{person}{Therese Quieta}.} \bibinfo{year}{2020}\natexlab{}.
\newblock \showarticletitle{Achieving Human Expert Level Time Performance for Atari Games--A Causal Learning Approach}. In \bibinfo{booktitle}{\emph{2020 IEEE Symposium Series on Computational Intelligence (SSCI)}}. IEEE, \bibinfo{pages}{449--456}.
\newblock


\bibitem[Holtz and Harold(2009)]%
        {holtz2009fair}
\bibfield{author}{\bibinfo{person}{Brian~C Holtz} {and} \bibinfo{person}{Crystal~M Harold}.} \bibinfo{year}{2009}\natexlab{}.
\newblock \showarticletitle{Fair today, fair tomorrow? A longitudinal investigation of overall justice perceptions.}
\newblock \bibinfo{journal}{\emph{Journal of applied psychology}} \bibinfo{volume}{94}, \bibinfo{number}{5} (\bibinfo{year}{2009}), \bibinfo{pages}{1185}.
\newblock


\bibitem[Howard and Borenstein(2018)]%
        {howard2018ugly}
\bibfield{author}{\bibinfo{person}{Ayanna Howard} {and} \bibinfo{person}{Jason Borenstein}.} \bibinfo{year}{2018}\natexlab{}.
\newblock \showarticletitle{The ugly truth about ourselves and our robot creations: the problem of bias and social inequity}.
\newblock \bibinfo{journal}{\emph{Science and engineering ethics}}  \bibinfo{volume}{24} (\bibinfo{year}{2018}), \bibinfo{pages}{1521--1536}.
\newblock


\bibitem[Howard and Kennedy~III(2020)]%
        {howard2020robots}
\bibfield{author}{\bibinfo{person}{Ayanna Howard} {and} \bibinfo{person}{Monroe Kennedy~III}.} \bibinfo{year}{2020}\natexlab{}.
\newblock \bibinfo{title}{Robots are not immune to bias and injustice}.
\newblock , \bibinfo{numpages}{eabf1364}~pages.
\newblock


\bibitem[Jiang et~al\mbox{.}(2021)]%
        {jiang2021generalized}
\bibfield{author}{\bibinfo{person}{Zhimeng Jiang}, \bibinfo{person}{Xiaotian Han}, \bibinfo{person}{Chao Fan}, \bibinfo{person}{Fan Yang}, \bibinfo{person}{Ali Mostafavi}, {and} \bibinfo{person}{Xia Hu}.} \bibinfo{year}{2021}\natexlab{}.
\newblock \showarticletitle{Generalized demographic parity for group fairness}. In \bibinfo{booktitle}{\emph{International Conference on Learning Representations}}.
\newblock


\bibitem[Jones and Skarlicki(2013)]%
        {jones2013perceptions}
\bibfield{author}{\bibinfo{person}{David~A Jones} {and} \bibinfo{person}{Daniel~P Skarlicki}.} \bibinfo{year}{2013}\natexlab{}.
\newblock \showarticletitle{How perceptions of fairness can change: A dynamic model of organizational justice}.
\newblock \bibinfo{journal}{\emph{Organizational psychology review}} \bibinfo{volume}{3}, \bibinfo{number}{2} (\bibinfo{year}{2013}), \bibinfo{pages}{138--160}.
\newblock


\bibitem[Jung et~al\mbox{.}(2020)]%
        {jung2020robot}
\bibfield{author}{\bibinfo{person}{Malte~F Jung}, \bibinfo{person}{Dominic DiFranzo}, \bibinfo{person}{Solace Shen}, \bibinfo{person}{Brett Stoll}, \bibinfo{person}{Houston Claure}, {and} \bibinfo{person}{Austin Lawrence}.} \bibinfo{year}{2020}\natexlab{}.
\newblock \showarticletitle{Robot-Assisted Tower Construction—A Method to Study the Impact of a Robot’s Allocation Behavior on Interpersonal Dynamics and Collaboration in Groups}.
\newblock \bibinfo{journal}{\emph{ACM Transactions on Human-Robot Interaction (THRI)}} \bibinfo{volume}{10}, \bibinfo{number}{1} (\bibinfo{year}{2020}), \bibinfo{pages}{1--23}.
\newblock


\bibitem[Kaminski(2014)]%
        {kaminski2014robots}
\bibfield{author}{\bibinfo{person}{Margot~E Kaminski}.} \bibinfo{year}{2014}\natexlab{}.
\newblock \showarticletitle{Robots in the home: What will we have agreed to}.
\newblock \bibinfo{journal}{\emph{Idaho L. Rev.}}  \bibinfo{volume}{51} (\bibinfo{year}{2014}), \bibinfo{pages}{661}.
\newblock


\bibitem[Kanda and Ishiguro(2005)]%
        {kanda2005communication}
\bibfield{author}{\bibinfo{person}{Takayuki Kanda} {and} \bibinfo{person}{Hiroshi Ishiguro}.} \bibinfo{year}{2005}\natexlab{}.
\newblock \showarticletitle{Communication robots for elementary schools}. In \bibinfo{booktitle}{\emph{Proceedings of the Symposium on Robot Companions: Hard Problems and Open Challenges in Robot-Human Interaction}}. The Society for the Study of Artificial Intelligence and the Simulation of~…, \bibinfo{pages}{54--63}.
\newblock


\bibitem[Kanda et~al\mbox{.}(2010)]%
        {kanda2010communication}
\bibfield{author}{\bibinfo{person}{Takayuki Kanda}, \bibinfo{person}{Masahiro Shiomi}, \bibinfo{person}{Zenta Miyashita}, \bibinfo{person}{Hiroshi Ishiguro}, {and} \bibinfo{person}{Norihiro Hagita}.} \bibinfo{year}{2010}\natexlab{}.
\newblock \showarticletitle{A communication robot in a shopping mall}.
\newblock \bibinfo{journal}{\emph{IEEE Transactions on Robotics}} \bibinfo{volume}{26}, \bibinfo{number}{5} (\bibinfo{year}{2010}), \bibinfo{pages}{897--913}.
\newblock


\bibitem[Keashly and Neuman(2010)]%
        {keashly2010faculty}
\bibfield{author}{\bibinfo{person}{Loraleigh Keashly} {and} \bibinfo{person}{Joel~H Neuman}.} \bibinfo{year}{2010}\natexlab{}.
\newblock \showarticletitle{Faculty experiences with bullying in higher education: Causes, consequences, and management}.
\newblock \bibinfo{journal}{\emph{Administrative Theory \& Praxis}} \bibinfo{volume}{32}, \bibinfo{number}{1} (\bibinfo{year}{2010}), \bibinfo{pages}{48--70}.
\newblock


\bibitem[Kim et~al\mbox{.}(2021)]%
        {kim2021age}
\bibfield{author}{\bibinfo{person}{Eugenia Kim}, \bibinfo{person}{De'Aira Bryant}, \bibinfo{person}{Deepak Srikanth}, {and} \bibinfo{person}{Ayanna Howard}.} \bibinfo{year}{2021}\natexlab{}.
\newblock \showarticletitle{Age bias in emotion detection: An analysis of facial emotion recognition performance on young, middle-aged, and older adults}. In \bibinfo{booktitle}{\emph{Proceedings of the 2021 AAAI/ACM Conference on AI, Ethics, and Society}}. \bibinfo{pages}{638--644}.
\newblock


\bibitem[Koenigs and Tranel(2007)]%
        {koenigs2007irrational}
\bibfield{author}{\bibinfo{person}{Michael Koenigs} {and} \bibinfo{person}{Daniel Tranel}.} \bibinfo{year}{2007}\natexlab{}.
\newblock \showarticletitle{Irrational economic decision-making after ventromedial prefrontal damage: evidence from the Ultimatum Game}.
\newblock \bibinfo{journal}{\emph{Journal of Neuroscience}} \bibinfo{volume}{27}, \bibinfo{number}{4} (\bibinfo{year}{2007}), \bibinfo{pages}{951--956}.
\newblock


\bibitem[Large et~al\mbox{.}(2020)]%
        {large2020studying}
\bibfield{author}{\bibinfo{person}{Jamie Large}, \bibinfo{person}{Graham Stodolski}, {and} \bibinfo{person}{Marynel V{\'a}zquez}.} \bibinfo{year}{2020}\natexlab{}.
\newblock \showarticletitle{Studying Human-Agent Interactions in Space Invaders}. In \bibinfo{booktitle}{\emph{Proceedings of the 8th International Conference on Human-Agent Interaction}}. \bibinfo{pages}{245--247}.
\newblock


\bibitem[Leventhal(1976)]%
        {leventhal1976distribution}
\bibfield{author}{\bibinfo{person}{Gerald~S Leventhal}.} \bibinfo{year}{1976}\natexlab{}.
\newblock \showarticletitle{The distribution of rewards and resources in groups and organizations}.
\newblock In \bibinfo{booktitle}{\emph{Advances in experimental social psychology}}. Vol.~\bibinfo{volume}{9}. \bibinfo{publisher}{Elsevier}, \bibinfo{pages}{91--131}.
\newblock


\bibitem[Lew et~al\mbox{.}(2023)]%
        {lew2023shutter}
\bibfield{author}{\bibinfo{person}{Alexander Lew}, \bibinfo{person}{Sydney Thompson}, \bibinfo{person}{Nathan Tsoi}, {and} \bibinfo{person}{Marynel V{\'a}zquez}.} \bibinfo{year}{2023}\natexlab{}.
\newblock \showarticletitle{Shutter, the Robot Photographer: Leveraging Behavior Trees for Public, In-the-Wild Human-Robot Interactions}.
\newblock \bibinfo{journal}{\emph{arXiv preprint arXiv:2302.00191}} (\bibinfo{year}{2023}).
\newblock


\bibitem[Litoiu et~al\mbox{.}(2015)]%
        {litoiu2015evidence}
\bibfield{author}{\bibinfo{person}{Alexandru Litoiu}, \bibinfo{person}{Daniel Ullman}, \bibinfo{person}{Jason Kim}, {and} \bibinfo{person}{Brian Scassellati}.} \bibinfo{year}{2015}\natexlab{}.
\newblock \showarticletitle{Evidence that robots trigger a cheating detector in humans}. In \bibinfo{booktitle}{\emph{Proceedings of the tenth annual acm/ieee international conference on human-robot interaction}}. \bibinfo{pages}{165--172}.
\newblock


\bibitem[Malle and Ullman(2021)]%
        {Malle2013}
\bibfield{author}{\bibinfo{person}{Bertram~F. Malle} {and} \bibinfo{person}{Daniel Ullman}.} \bibinfo{year}{2021}\natexlab{}.
\newblock \showarticletitle{Chapter 1 - A multidimensional conception and measure of human-robot trust}.
\newblock In \bibinfo{booktitle}{\emph{Trust in Human-Robot Interaction}}, \bibfield{editor}{\bibinfo{person}{Chang~S. Nam} {and} \bibinfo{person}{Joseph~B. Lyons}} (Eds.). \bibinfo{publisher}{Academic Press}, \bibinfo{pages}{3--25}.
\newblock
\showISBNx{978-0-12-819472-0}
\urldef\tempurl%
\url{https://doi.org/10.1016/B978-0-12-819472-0.00001-0}
\showDOI{\tempurl}


\bibitem[Mateas(2003)]%
        {mateas2003expressive}
\bibfield{author}{\bibinfo{person}{Michael Mateas}.} \bibinfo{year}{2003}\natexlab{}.
\newblock \showarticletitle{Expressive AI: Games and Artificial Intelligence.}. In \bibinfo{booktitle}{\emph{DiGRA Conference}}, Vol.~\bibinfo{volume}{15}. Citeseer.
\newblock


\bibitem[Matta et~al\mbox{.}(2017)]%
        {matta2017consistently}
\bibfield{author}{\bibinfo{person}{Fadel~K Matta}, \bibinfo{person}{Brent~A Scott}, \bibinfo{person}{Jason~A Colquitt}, \bibinfo{person}{Joel Koopman}, {and} \bibinfo{person}{Liana~G Passantino}.} \bibinfo{year}{2017}\natexlab{}.
\newblock \showarticletitle{Is consistently unfair better than sporadically fair? An investigation of justice variability and stress}.
\newblock \bibinfo{journal}{\emph{Academy of Management Journal}} \bibinfo{volume}{60}, \bibinfo{number}{2} (\bibinfo{year}{2017}), \bibinfo{pages}{743--770}.
\newblock


\bibitem[Mitchell et~al\mbox{.}(2021)]%
        {mitchell2021algorithmic}
\bibfield{author}{\bibinfo{person}{Shira Mitchell}, \bibinfo{person}{Eric Potash}, \bibinfo{person}{Solon Barocas}, \bibinfo{person}{Alexander D'Amour}, {and} \bibinfo{person}{Kristian Lum}.} \bibinfo{year}{2021}\natexlab{}.
\newblock \showarticletitle{Algorithmic fairness: Choices, assumptions, and definitions}.
\newblock \bibinfo{journal}{\emph{Annual Review of Statistics and Its Application}}  \bibinfo{volume}{8} (\bibinfo{year}{2021}), \bibinfo{pages}{141--163}.
\newblock


\bibitem[Moore(2009)]%
        {moore2009fairness}
\bibfield{author}{\bibinfo{person}{Chris Moore}.} \bibinfo{year}{2009}\natexlab{}.
\newblock \showarticletitle{Fairness in children's resource allocation depends on the recipient}.
\newblock \bibinfo{journal}{\emph{Psychological Science}} \bibinfo{volume}{20}, \bibinfo{number}{8} (\bibinfo{year}{2009}), \bibinfo{pages}{944--948}.
\newblock


\bibitem[Mutlu et~al\mbox{.}(2009)]%
        {mutlu2009footing}
\bibfield{author}{\bibinfo{person}{Bilge Mutlu}, \bibinfo{person}{Toshiyuki Shiwa}, \bibinfo{person}{Takayuki Kanda}, \bibinfo{person}{Hiroshi Ishiguro}, {and} \bibinfo{person}{Norihiro Hagita}.} \bibinfo{year}{2009}\natexlab{}.
\newblock \showarticletitle{Footing in human-robot conversations: how robots might shape participant roles using gaze cues}. In \bibinfo{booktitle}{\emph{Proceedings of the 4th ACM/IEEE international conference on Human robot interaction}}. \bibinfo{pages}{61--68}.
\newblock


\bibitem[Nashed et~al\mbox{.}(2023)]%
        {nashed2023fairness}
\bibfield{author}{\bibinfo{person}{Samer~B Nashed}, \bibinfo{person}{Justin Svegliato}, {and} \bibinfo{person}{Su~Lin Blodgett}.} \bibinfo{year}{2023}\natexlab{}.
\newblock \showarticletitle{Fairness and Sequential Decision Making: Limits, Lessons, and Opportunities}.
\newblock \bibinfo{journal}{\emph{arXiv preprint arXiv:2301.05753}} (\bibinfo{year}{2023}).
\newblock


\bibitem[Nicklin et~al\mbox{.}(2011)]%
        {nicklin2011importance}
\bibfield{author}{\bibinfo{person}{Jessica~M Nicklin}, \bibinfo{person}{Rebecca Greenbaum}, \bibinfo{person}{Laurel~A McNall}, \bibinfo{person}{Robert Folger}, {and} \bibinfo{person}{Kevin~J Williams}.} \bibinfo{year}{2011}\natexlab{}.
\newblock \showarticletitle{The importance of contextual variables when judging fairness: An examination of counterfactual thoughts and fairness theory}.
\newblock \bibinfo{journal}{\emph{Organizational Behavior and Human Decision Processes}} \bibinfo{volume}{114}, \bibinfo{number}{2} (\bibinfo{year}{2011}), \bibinfo{pages}{127--141}.
\newblock


\bibitem[Ogunyale et~al\mbox{.}(2018)]%
        {ogunyale2018does}
\bibfield{author}{\bibinfo{person}{Tobi Ogunyale}, \bibinfo{person}{De'Aira Bryant}, {and} \bibinfo{person}{Ayanna Howard}.} \bibinfo{year}{2018}\natexlab{}.
\newblock \showarticletitle{Does removing stereotype priming remove bias? A pilot human-robot interaction study}.
\newblock \bibinfo{journal}{\emph{arXiv preprint arXiv:1807.00948}} (\bibinfo{year}{2018}).
\newblock


\bibitem[Pessach and Shmueli(2020)]%
        {pessach2020algorithmic}
\bibfield{author}{\bibinfo{person}{Dana Pessach} {and} \bibinfo{person}{Erez Shmueli}.} \bibinfo{year}{2020}\natexlab{}.
\newblock \showarticletitle{Algorithmic fairness}.
\newblock \bibinfo{journal}{\emph{arXiv preprint arXiv:2001.09784}} (\bibinfo{year}{2020}).
\newblock


\bibitem[Ranz et~al\mbox{.}(2017)]%
        {ranz2017capability}
\bibfield{author}{\bibinfo{person}{Fabian Ranz}, \bibinfo{person}{Vera Hummel}, {and} \bibinfo{person}{Wilfried Sihn}.} \bibinfo{year}{2017}\natexlab{}.
\newblock \showarticletitle{Capability-based task allocation in human-robot collaboration}.
\newblock \bibinfo{journal}{\emph{Procedia Manufacturing}}  \bibinfo{volume}{9} (\bibinfo{year}{2017}), \bibinfo{pages}{182--189}.
\newblock


\bibitem[Sakamoto and Ono(2006)]%
        {sakamoto2006sociality}
\bibfield{author}{\bibinfo{person}{Daisuke Sakamoto} {and} \bibinfo{person}{Tetsuo Ono}.} \bibinfo{year}{2006}\natexlab{}.
\newblock \showarticletitle{Sociality of robots: do robots construct or collapse human relations?}. In \bibinfo{booktitle}{\emph{Proceedings of the 1st ACM SIGCHI/SIGART conference on human-robot interaction}}. \bibinfo{pages}{355--356}.
\newblock


\bibitem[Sanfey et~al\mbox{.}(2003)]%
        {sanfey2003neural}
\bibfield{author}{\bibinfo{person}{Alan~G Sanfey}, \bibinfo{person}{James~K Rilling}, \bibinfo{person}{Jessica~A Aronson}, \bibinfo{person}{Leigh~E Nystrom}, {and} \bibinfo{person}{Jonathan~D Cohen}.} \bibinfo{year}{2003}\natexlab{}.
\newblock \showarticletitle{The neural basis of economic decision-making in the ultimatum game}.
\newblock \bibinfo{journal}{\emph{Science}} \bibinfo{volume}{300}, \bibinfo{number}{5626} (\bibinfo{year}{2003}), \bibinfo{pages}{1755--1758}.
\newblock


\bibitem[Saxena et~al\mbox{.}(2019)]%
        {saxena2019fairness}
\bibfield{author}{\bibinfo{person}{Nripsuta~Ani Saxena}, \bibinfo{person}{Karen Huang}, \bibinfo{person}{Evan DeFilippis}, \bibinfo{person}{Goran Radanovic}, \bibinfo{person}{David~C Parkes}, {and} \bibinfo{person}{Yang Liu}.} \bibinfo{year}{2019}\natexlab{}.
\newblock \showarticletitle{How do fairness definitions fare? Examining public attitudes towards algorithmic definitions of fairness}. In \bibinfo{booktitle}{\emph{Proceedings of the 2019 AAAI/ACM Conference on AI, Ethics, and Society}}. \bibinfo{pages}{99--106}.
\newblock


\bibitem[Sebo et~al\mbox{.}(2020)]%
        {sebo2020robots}
\bibfield{author}{\bibinfo{person}{Sarah Sebo}, \bibinfo{person}{Brett Stoll}, \bibinfo{person}{Brian Scassellati}, {and} \bibinfo{person}{Malte~F Jung}.} \bibinfo{year}{2020}\natexlab{}.
\newblock \showarticletitle{Robots in groups and teams: a literature review}.
\newblock \bibinfo{journal}{\emph{Proceedings of the ACM on Human-Computer Interaction}} \bibinfo{volume}{4}, \bibinfo{number}{CSCW2} (\bibinfo{year}{2020}), \bibinfo{pages}{1--36}.
\newblock


\bibitem[Shah et~al\mbox{.}(2011)]%
        {shah2011improved}
\bibfield{author}{\bibinfo{person}{Julie Shah}, \bibinfo{person}{James Wiken}, \bibinfo{person}{Brian Williams}, {and} \bibinfo{person}{Cynthia Breazeal}.} \bibinfo{year}{2011}\natexlab{}.
\newblock \showarticletitle{Improved human-robot team performance using chaski, a human-inspired plan execution system}. In \bibinfo{booktitle}{\emph{Proceedings of the 6th international conference on Human-robot interaction}}. ACM, \bibinfo{pages}{29--36}.
\newblock


\bibitem[Short et~al\mbox{.}(2010)]%
        {short2010no}
\bibfield{author}{\bibinfo{person}{Elaine Short}, \bibinfo{person}{Justin Hart}, \bibinfo{person}{Michelle Vu}, {and} \bibinfo{person}{Brian Scassellati}.} \bibinfo{year}{2010}\natexlab{}.
\newblock \showarticletitle{No fair!! an interaction with a cheating robot}. In \bibinfo{booktitle}{\emph{2010 5th ACM/IEEE International Conference on Human-Robot Interaction (HRI)}}. IEEE, \bibinfo{pages}{219--226}.
\newblock


\bibitem[Steiner and Rain(1989)]%
        {steiner1989immediate}
\bibfield{author}{\bibinfo{person}{Dirk~D Steiner} {and} \bibinfo{person}{Jeffrey~S Rain}.} \bibinfo{year}{1989}\natexlab{}.
\newblock \showarticletitle{Immediate and delayed primacy and recency effects in performance evaluation.}
\newblock \bibinfo{journal}{\emph{Journal of Applied Psychology}} \bibinfo{volume}{74}, \bibinfo{number}{1} (\bibinfo{year}{1989}), \bibinfo{pages}{136}.
\newblock


\bibitem[Suzuki et~al\mbox{.}(2022)]%
        {suzuki2022augmented}
\bibfield{author}{\bibinfo{person}{Ryo Suzuki}, \bibinfo{person}{Adnan Karim}, \bibinfo{person}{Tian Xia}, \bibinfo{person}{Hooman Hedayati}, {and} \bibinfo{person}{Nicolai Marquardt}.} \bibinfo{year}{2022}\natexlab{}.
\newblock \showarticletitle{Augmented reality and robotics: A survey and taxonomy for ar-enhanced human-robot interaction and robotic interfaces}. In \bibinfo{booktitle}{\emph{Proceedings of the 2022 CHI Conference on Human Factors in Computing Systems}}. \bibinfo{pages}{1--33}.
\newblock


\bibitem[Tennent et~al\mbox{.}(2019)]%
        {tennent2019micbot}
\bibfield{author}{\bibinfo{person}{Hamish Tennent}, \bibinfo{person}{Solace Shen}, {and} \bibinfo{person}{Malte Jung}.} \bibinfo{year}{2019}\natexlab{}.
\newblock \showarticletitle{Micbot: A peripheral robotic object to shape conversational dynamics and team performance}. In \bibinfo{booktitle}{\emph{2019 14th ACM/IEEE International Conference on Human-Robot Interaction (HRI)}}. IEEE, \bibinfo{pages}{133--142}.
\newblock


\bibitem[Thompson et~al\mbox{.}(2024)]%
        {thompson2024shutter}
\bibfield{author}{\bibinfo{person}{Sydney Thompson}, \bibinfo{person}{Austin Narcomey}, \bibinfo{person}{Alexander Lew}, {and} \bibinfo{person}{Marynel V{\'a}zquez}.} \bibinfo{year}{2024}\natexlab{}.
\newblock \showarticletitle{Shutter: A Low-Cost and Flexible Social Robot Platform for In-the-Wild Deployments}. In \bibinfo{booktitle}{\emph{Companion of the 2024 ACM/IEEE International Conference on Human-Robot Interaction}}. \bibinfo{pages}{94--96}.
\newblock


\bibitem[Tolmeijer et~al\mbox{.}(2022)]%
        {Tolmeijer2022}
\bibfield{author}{\bibinfo{person}{Suzanne Tolmeijer}, \bibinfo{person}{Markus Christen}, \bibinfo{person}{Serhiy Kandul}, \bibinfo{person}{Markus Kneer}, {and} \bibinfo{person}{Abraham Bernstein}.} \bibinfo{year}{2022}\natexlab{}.
\newblock \showarticletitle{Capable but Amoral? Comparing AI and Human Expert Collaboration in Ethical Decision Making}. In \bibinfo{booktitle}{\emph{Proceedings of the 2022 CHI Conference on Human Factors in Computing Systems}} (<conf-loc>, <city>New Orleans</city>, <state>LA</state>, <country>USA</country>, </conf-loc>) \emph{(\bibinfo{series}{CHI '22})}. \bibinfo{publisher}{Association for Computing Machinery}, \bibinfo{address}{New York, NY, USA}, Article \bibinfo{articleno}{160}, \bibinfo{numpages}{17}~pages.
\newblock
\showISBNx{9781450391573}
\urldef\tempurl%
\url{https://doi.org/10.1145/3491102.3517732}
\showDOI{\tempurl}


\bibitem[Turvey and Freeman(2012)]%
        {TURVEY2012495}
\bibfield{author}{\bibinfo{person}{B.E. Turvey} {and} \bibinfo{person}{J.L. Freeman}.} \bibinfo{year}{2012}\natexlab{}.
\newblock \showarticletitle{Jury Psychology}.
\newblock In \bibinfo{booktitle}{\emph{Encyclopedia of Human Behavior (Second Edition)} (\bibinfo{edition}{second edition} ed.)}, \bibfield{editor}{\bibinfo{person}{V.S. Ramachandran}} (Ed.). \bibinfo{publisher}{Academic Press}, \bibinfo{address}{San Diego}, \bibinfo{pages}{495--502}.
\newblock


\bibitem[Uhde et~al\mbox{.}(2020)]%
        {uhde2020fairness}
\bibfield{author}{\bibinfo{person}{Alarith Uhde}, \bibinfo{person}{Nadine Schlicker}, \bibinfo{person}{Dieter~P Wallach}, {and} \bibinfo{person}{Marc Hassenzahl}.} \bibinfo{year}{2020}\natexlab{}.
\newblock \showarticletitle{Fairness and decision-making in collaborative shift scheduling systems}. In \bibinfo{booktitle}{\emph{Proceedings of the 2020 CHI Conference on Human Factors in Computing Systems}}. \bibinfo{pages}{1--13}.
\newblock


\bibitem[Ullman and Malle(2020)]%
        {mdmtv2}
\bibfield{author}{\bibinfo{person}{Daniel Ullman} {and} \bibinfo{person}{Bertram~F. Malle}.} \bibinfo{year}{2020}\natexlab{}.
\newblock \bibinfo{title}{{MDMT: Multi-Dimensional Measure of Trust}}.
\newblock
\newblock
\urldef\tempurl%
\url{https://research.clps.brown.edu/SocCogSci/Measures/MDMT_v2.pdf}
\showURL{%
\tempurl}


\bibitem[Van~der Hoorn et~al\mbox{.}(2021)]%
        {van2021think}
\bibfield{author}{\bibinfo{person}{Diede~PM Van~der Hoorn}, \bibinfo{person}{Anouk Neerincx}, {and} \bibinfo{person}{Maartje~MA de Graaf}.} \bibinfo{year}{2021}\natexlab{}.
\newblock \showarticletitle{" I think you are doing a bad job!" The Effect of Blame Attribution by a Robot in Human-Robot Collaboration}. In \bibinfo{booktitle}{\emph{Proceedings of the 2021 ACM/IEEE international conference on human-robot interaction}}. \bibinfo{pages}{140--148}.
\newblock


\bibitem[Wilson et~al\mbox{.}(2019)]%
        {wilson2019predictive}
\bibfield{author}{\bibinfo{person}{Benjamin Wilson}, \bibinfo{person}{Judy Hoffman}, {and} \bibinfo{person}{Jamie Morgenstern}.} \bibinfo{year}{2019}\natexlab{}.
\newblock \showarticletitle{Predictive inequity in object detection}.
\newblock \bibinfo{journal}{\emph{arXiv preprint arXiv:1902.11097}} (\bibinfo{year}{2019}).
\newblock


\bibitem[Yang(2021)]%
        {yang2021toward}
\bibfield{author}{\bibinfo{person}{Qiang Yang}.} \bibinfo{year}{2021}\natexlab{}.
\newblock \showarticletitle{Toward responsible ai: An overview of federated learning for user-centered privacy-preserving computing}.
\newblock \bibinfo{journal}{\emph{ACM Transactions on Interactive Intelligent Systems (TiiS)}} \bibinfo{volume}{11}, \bibinfo{number}{3-4} (\bibinfo{year}{2021}), \bibinfo{pages}{1--22}.
\newblock


\bibitem[Yang et~al\mbox{.}(2017)]%
        {yang2017evaluating}
\bibfield{author}{\bibinfo{person}{X~Jessie Yang}, \bibinfo{person}{Vaibhav~V Unhelkar}, \bibinfo{person}{Kevin Li}, {and} \bibinfo{person}{Julie~A Shah}.} \bibinfo{year}{2017}\natexlab{}.
\newblock \showarticletitle{Evaluating effects of user experience and system transparency on trust in automation}. In \bibinfo{booktitle}{\emph{Proceedings of the 2017 ACM/IEEE international conference on human-robot interaction}}. \bibinfo{pages}{408--416}.
\newblock


\bibitem[Yew(2021)]%
        {yew2021trust}
\bibfield{author}{\bibinfo{person}{Gary Chan~Kok Yew}.} \bibinfo{year}{2021}\natexlab{}.
\newblock \showarticletitle{Trust in and ethical design of carebots: the case for ethics of care}.
\newblock \bibinfo{journal}{\emph{International Journal of Social Robotics}} \bibinfo{volume}{13}, \bibinfo{number}{4} (\bibinfo{year}{2021}), \bibinfo{pages}{629--645}.
\newblock


\bibitem[You et~al\mbox{.}(2011)]%
        {you2011robot}
\bibfield{author}{\bibinfo{person}{Sangseok You}, \bibinfo{person}{Jiaqi Nie}, \bibinfo{person}{Kiseul Suh}, {and} \bibinfo{person}{S~Shyam Sundar}.} \bibinfo{year}{2011}\natexlab{}.
\newblock \showarticletitle{When the robot criticizes you... Self-serving bias in human-robot interaction}. In \bibinfo{booktitle}{\emph{Proceedings of the 6th international conference on human-robot interaction}}. \bibinfo{pages}{295--296}.
\newblock


\bibitem[Zhang et~al\mbox{.}(2023)]%
        {zhang2023self}
\bibfield{author}{\bibinfo{person}{Qiping Zhang}, \bibinfo{person}{Austin Narcomey}, \bibinfo{person}{Kate Candon}, {and} \bibinfo{person}{Marynel V{\'a}zquez}.} \bibinfo{year}{2023}\natexlab{}.
\newblock \showarticletitle{Self-Annotation Methods for Aligning Implicit and Explicit Human Feedback in Human-Robot Interaction}. In \bibinfo{booktitle}{\emph{Proceedings of the 2023 ACM/IEEE International Conference on Human-Robot Interaction}}. \bibinfo{pages}{398--407}.
\newblock


\bibitem[Zhou and Brandao({[n.\,d.]})]%
        {zhounoise}
\bibfield{author}{\bibinfo{person}{Zewei Zhou} {and} \bibinfo{person}{Martim Brandao}.} \bibinfo{year}{[n.\,d.]}\natexlab{}.
\newblock \showarticletitle{Noise and Environmental Justice in Drone Fleet Delivery Paths: A Simulation-Based Audit and Algorithm for Fairer Impact Distribution}.
\newblock  (\bibinfo{year}{[n.\,d.]}).
\newblock


\end{thebibliography}

\end{document}